\documentclass[conference,compsoc]{IEEEtran}
\usepackage{verbatim}
\usepackage{graphicx,  amsbsy, cuted}
\usepackage{subfigure} 
\usepackage{xcolor}
\usepackage{multirow}
\usepackage{bbm}
\usepackage{booktabs}
\usepackage{threeparttable}
\usepackage{diagbox}
\usepackage{algorithmic}
\usepackage{amsmath}
\usepackage[ruled]{algorithm2e}
\usepackage{amsfonts}
\usepackage{url}

\begin{document}

\title{GRID: Protecting Training Graph from Link Stealing Attacks on GNN Models}

\author{
    \IEEEauthorblockN{Jiadong Lou\IEEEauthorrefmark{1}, Xu Yuan\IEEEauthorrefmark{1}, Rui Zhang\IEEEauthorrefmark{1}, Xingliang Yuan\IEEEauthorrefmark{2}, Neil Zhenqiang Gong\IEEEauthorrefmark{3}, and Nian-Feng Tzeng\IEEEauthorrefmark{4}}
    \IEEEauthorblockA{\IEEEauthorrefmark{1}University of Delaware,  \IEEEauthorrefmark{2}The University of Melbourne, }
    \IEEEauthorblockA{\IEEEauthorrefmark{3}Duke University,  \IEEEauthorrefmark{4}University of Louisiana at Lafayette}
}

\maketitle
\pagestyle{empty}
\thispagestyle{empty}
\begin{abstract}

Graph neural networks (GNNs) have exhibited superior performance in various classification tasks on graph-structured data.  
However, they encounter the potential vulnerability from the link stealing attacks, which can infer the presence of a link between two nodes via measuring the similarity of its incident nodes' prediction vectors produced by a GNN model. 
Such attacks pose severe security and privacy threats to the training graph used in GNN models. 
In this work, we propose a novel solution, called Graph Link Disguise (GRID), to defend against link stealing attacks with the formal guarantee of GNN model utility for retaining prediction accuracy.
The key idea of GRID is to add carefully crafted noises to the nodes' prediction vectors for disguising adjacent nodes as n-hop indirect neighboring nodes.
We take into account the graph topology and select only a subset of nodes (called \emph{core nodes}) covering all links for adding noises, which can avert the noises offset and have the further advantages of reducing both the distortion loss and the computation cost.
Our crafted noises can ensure 1) the noisy prediction vectors of any two adjacent nodes have their similarity level like that of two non-adjacent nodes and 2) the model prediction is unchanged to ensure zero utility loss. 
{Extensive experiments on five datasets are conducted to show the effectiveness of our proposed GRID solution against different representative link-stealing attacks under transductive settings and inductive settings respectively, as well as two influence-based attacks. }
Meanwhile, it achieves a much better privacy-utility trade-off than existing methods when extended to GNNs.
\end{abstract}


\section{Introduction}
\vspace{-0.5em}
Graph-structured data, where nodes represent entities and edges describe their relationships, are prevalent in various real-world applications, such as social networks, molecular graphs, and natural language processing. 
To handle such data, Graph Neural Networks (GNNs) \cite{scarselli2008graph,hamilton2017inductive, velivckovic2018graph} were developed to effectively learn node features and capture the complex interactions among nodes, enabling better performance in tasks like node classification and link prediction.
Despite the rapid advancement of GNN models, significant security and privacy concerns have surfaced, particularly regarding sensitive information leakage within the graph topology, such as link interactions. 
Accordingly, link stealing attacks were introduced in \cite{he2021stealing,wu2022linkteller, meng2023devil,wu2024link,zhang2023demystifying}, aiming to extract the topology information of a training graph by inferring the presence of a link between any two queried nodes.  
The intuition behind link inference attacks stems from the fact that the message-passing mechanism of GNNs aggregates information from neighboring nodes for each node, resulting in distinctive patterns in the predictions of neighboring nodes, such as high similarity.
Such attacks, which extract links in the training graph of target GNN models, can lead to user privacy leakage, intellectual property infringement, among others.

Witnessing such a link leakage risk, it is unfortunate that existing defensive solutions designed to thwart machine learning inference attacks are unsuitable for defending against link-stealing attacks.
These defensive methods can be broadly grouped into three categories: 
1) by adding additional terms in loss functions to obscure training dataset patterns~\cite{shokri2017membership,nasr2018machine}; 
2) by adding crafted perturbations to the training process via differential privacy techniques~\cite{shokri2015privacy,abadi2016deep,wang2017differentially,yu2019differentially,wu2022linkteller,sajadmanesh2023gap}; 3) by adding noises to the prediction vectors of the trained model~\cite{jia2019memguard}. 
The first two categories of solutions will interfere with the model training process, inevitably hurting model performance to some extent. 
For example, the pioneering differential privacy method as reported in~\cite{sajadmanesh2023gap} incurs over 20\% model accuracy degradation when defending against the node inference attacks. 
The third category of solution aims to craft the noise independently for each prediction vector, thereby failing to consider the similarity patterns among adjacent nodes governed by the GNN model and the graph structure.
Hence, it is urgent to call for a new defensive scheme that can take into account both the graph topology and GNN information aggregation characteristics to protect the GNN models from link-stealing attacks while formally ensuring model utility guarantees.

{This paper presents a novel defense method, named {\bf Gr}aph L{\bf i}nk {\bf D}isguise ({\bf GRID}), to prevent link-stealing attacks against GNN models, being the first to provide a formal guarantee for model utility.}
GRID adds noise vectors to the prediction vectors of nodes in the training graph to distort the similarity of adjacent nodes, thereby obscuring the existence of links. 
We formulate this defense scheme as an optimization problem to achieve two goals: 1) preventing the attacker's link-stealing classifiers from determining if a link exists between two nodes based on their queried prediction vectors, and 2) maintaining the accuracy of the model's prediction results.
However, a number of challenges arise in solving the noise generation problem. 
First, the inherent data aggregation among neighboring nodes in GNN models makes it difficult to dissociate the correlation between similar predictions and link presence, while only slight noises are tolerant for retaining model predictions. 
Second, given the large and complex real-world graph topologies, simply adding noise to the prediction vectors of all nodes results in unacceptable computational complexity and noise offset. 
Therefore, to avert conflicts and improve computational efficiency, graph topology information should be considered when crafting the noise vectors, which represents another huge challenge.

To tackle the aforementioned challenges, we propose a novel solution with the key idea of disguising adjacent nodes as n-hop indirect neighboring nodes and crafting noise only for selected core nodes' prediction vectors.
First, instead of minimizing the prediction similarity of adjacent nodes, we disguise them by reducing their similarity patterns to the level of indirectly connected neighboring nodes. 
This approach is sufficient to disrupt link inference and thwart adaptive attacks, as adjacent nodes' prediction vectors will no longer exhibit distinctive similarity patterns.
Additionally, considering the interconnections among nodes, we identify a subset of nodes (dubbed core nodes) that cover all links in the training graph. 
By targeting noise additions only to core nodes, we can avoid conflicts among nodes and reduce computational complexity.
Thus, we reformulate our defense problem by minimizing the similarity gap between adjacent nodes and indirectly connected neighboring nodes, calculating the necessary noise for core nodes accordingly.

{We conduct extensive experiments on five graph datasets (comprising three citation datasets and two chemical datasets) and different GNNs (both transductive and inductive models) under representative link-stealing attacks proposed in \cite{he2021stealing,wu2024link} to evaluate the effectiveness of our GRID method. }
Our experimental results demonstrate that GRID can substantially degrade the inference performance of all these link-stealing attacks, reducing their accuracy to around 50\% (similar to random guessing) with appropriate parameter settings.
Additionally, we compare our GRID with five state-of-the-art defense approaches across three categories: employing regularization terms, applying differential privacy techniques, and adding prediction noise \cite{shokri2017membership, srivastava2014dropout, nasr2018machine, jia2019memguard,sajadmanesh2023gap}. The comparative results show that our GRID significantly outperforms all five counterparts in terms of the defense-utility trade-off.

Our contributions can be summarized as follows
\begin{itemize}
\vspace{-0.3em}
    \item We propose the first Graph Link Disguise method, i.e., GRID, to prevent link stealing attacks on GNN models with the formal utility guarantee, serving as a new framework for protecting the graph topology information. 
    \item We propose a novel scheme of crafting noises for a subset of nodes to disguise the adjacent nodes into n-hop indirect neighboring nodes, able to defend against link inference attacks.
    \item We conduct comprehensive experiments to demonstrate that our GRID effectively defends against link-stealing attacks and outperforms all existing methods in terms of the defense-utility trade-off. 
\end{itemize}


\thispagestyle{empty}
\vspace{-0.5em}
\section{Background}
\vspace{-0.5em}
\label{sec:Preliminary}

\subsection{Graph Neural Networks}
\vspace{-0.5em}
Graph neural networks (GNNs) are neural network techniques tailored for learning from graph-structured data. 
Typically, a node classification GNN model, denoted as $f_g()$, takes the graph topology with a set of labeled nodes and their attributes as its input for training.
This model is then used to predict the labels of remaining unlabeled nodes. 
That is, the training dataset can be expressed as $\mathcal{G}=(\mathcal{N}, \mathcal{A}^{\mathcal{N}},\mathcal{F},\mathcal{L})$, where $\mathcal{N}$ denotes a set of nodes, $\mathcal{A}^{\mathcal{N}}$  represents the adjacency matrix which depicts the graph topology, $\mathcal{F}$ represents  node attributes, and  $\mathcal{L}$  indicates labels. 
The node classification model $f_g()$  takes  $\mathcal{G}$ for training and then predicts the labels of unlabeled nodes.

GNNs can be categorized as transductive and inductive learning paradigms. 
In transductive learning, GNNs are trained on both labeled and unlabeled nodes of a fixed graph and then used to predict the labels of the unlabeled nodes within this graph.
It is typically used in scenarios where the graph structure is fixed, such as predicting protein functions or gene associations within the detected protein-protein interaction or gene regulatory graphs, with Graph Convolutional Networks (GCNs) being the classic model proposed for this paradigm.
In inductive learning, GNNs are trained on labeled nodes and can predict the labels of nodes unseen during the training process, with Graph Attention Networks (GATs)~\cite{velivckovic2018graph}, Graph Transformer Networks (GTNs)~\cite{yun2019graph}, among others being proposed.
Inductive GNNs are used in dynamic scenarios where new nodes or edges can appear, and the model needs to generalize to unseen data, such as providing recommendations for new users in recommendation systems and analyzing new users and their potential connections in a dynamic social network.

\vspace{-0.5em}
\subsection{Link Stealing Attacks on GNNs} 
\vspace{-0.5em}
\label{Subsec:link steal}

The link inference of GNN models was first proposed in~\cite{he2021stealing}, called the link stealing attacks, which aim to infer the presence of a link between two nodes in the training graph. 
In these attacks, an attacker can query data samples over a trained GNN model, to obtain the corresponding prediction vectors. 
Based on this,  the attacker attempts to infer whether a link exists between the queried nodes $n_i$ and $n_j$ in the training graph.
The intuition of different kinds of link inference attacks stems from the same fact that GNNs aggregate information for each node from its neighboring nodes.
According to this inherent GNN nature, two lines of attacks are proposed.

\subsubsection{Similarity-based Attack} 
In~\cite{he2021stealing}, authors first propose to exploit the characteristic in the prediction vectors of two neighboring nodes that if two nodes are linked, their prediction vectors queried from the target model should be similar, as they have shared some similar information. 
Therefore, the attacker can learn the similarity difference between the prediction vectors of linked or unlinked node pairs to infer whether there is a link between them.
Mathematically, given two nodes $i$ and $j$, the attacker aims at training a link-stealing classifier, denoted as  $f_m()$, to have
\begin{equation}
	\label{classifer2}
	f_m\big(\mathcal{E}({\bf v_i},{\bf v_j})\big)=
	\begin{cases}
		1& \text{if } \mathcal{A}_{ij}=1\ ;\\
		0& \text{otherwise}\ ,\\ 
	\end{cases}
\end{equation}
where ${\bf v_i}=f_g(n_i)$, ${\bf v_j}=f_g(n_j)$, representing their prediction vectors, and $\mathcal{E}()$ captures the similarity metrics. 
$\mathcal{A}_{ij}=1$ represents there is a link between $n_i$ and $n_j$.
In practice, the link-stealing classifier's output, i.e., $f_m()$, is a confidence score in the range of  0 to 1. 
If it is greater than $0.5$, the classifier is confident to predict that there is a link between the two queried nodes; otherwise, no link exists. 

{
In~\cite{he2021stealing}, eight types of attacks (defined as Attack-0 to Attack-7) were proposed as shown in Table~\ref{tab:attacks}, assuming an attacker has different levels of background knowledge, including 1) partial node attributes from the target dataset, 2) partial graph topology of the target dataset and 3) a shadow dataset.
Among these, Attack-0 and Attack-1 are two representative attacks that cover unsupervised and supervised black-box scenarios, while Attack-6 is the strongest gray-box attack, where the attacker can access partial graph topology and node attributes.
Although initially designed for transductive models, these attack schemes have also been shown to be effective on inductive GNN models, as both models output predictions for nodes in the training graph regardless of whether they are applied in transductive or inductive tasks.
For example, attacks proposed in~\cite{wu2024link}, abbreviated as ILS, extend these ideas to inductive GNN models by calculating and differentiating metrics between the prediction vectors of pairwise nodes. 
Ten attacks were proposed to cover various scenarios, with Attack A2 and A9 being reported as the strongest in the posterior-only and combined attack categories, respectively.
These similarity-based link-stealing attacks are both general and practical, posing a significant threat to the training graphs.}

\begin{table}[t]
	\footnotesize
	\centering
	\caption{Eight attacks and their background knowledge}
	\vspace{-0.5em}
	\begin{tabular}{c|ccc|c|ccc}
		\toprule
		Attack   & $\mathcal{F}$ &  $\mathcal{A}^*$ & $\mathcal{D}'$ & Attack   &  $\mathcal{F}$ & $\mathcal{A}^*$ & $\mathcal{D}'$ \\
		\midrule
		Attack-0 & $\times$  & $\times$  & $\times$  & Attack-4 & $\times$  & $\checkmark$  & $\checkmark$  \\
		Attack-1 &  $\times$ & $\times$  & $\checkmark$  & Attack-5 & $\checkmark$  & $\times$  & $\checkmark$  \\
		Attack-2 & $\checkmark$  &  $\times$ &$\times$   & Attack-6 & $\checkmark$  & $\checkmark$  & $\times$  \\
		Attack-3 & $\times$  & $\checkmark$  & $\times$  & Attack-7 & $\checkmark$  & $\checkmark$  & $\checkmark$ \\
		\bottomrule
	\end{tabular}
	\label{tab:attacks}
	 \vspace{-1.5em}
\end{table}

\subsubsection{Influence-based Attack} 
The second type of link-stealing attack was proposed in~\cite{wu2022linkteller, meng2023devil}, which leverages the observation that two nodes are more likely to be linked if changing the features of one node significantly influences the prediction results of the other node due to the message-passing mechanism. 
To effectively capture such influence delivered through the link, the authors introduce the graph data poisoning scenario where an attacker can perform malicious node injection (also called infiltration) into the training graph or maliciously change the attributes of the controlled nodes during the inference stage.

Taking the latest attacks proposed in~\cite{meng2023devil} as an example, the attacker aims at inferring the link between two target nodes $u$ and $v$.
The attacker first injects a node $a$ and a link $(a,u)$ to the training graph $\mathcal{G}$. 
Then it queries the target model to obtain the prediction for node $a$, denoted as $f_{g}(a)$.
Second, the attacker injects another node $b$ and a link $(b,v)$ to the training graph $\mathcal{G}$ and again queries the target model to obtain the prediction for node $a$, denoted as $f_{g}(a|b)$.
If $u$ and $v$ are linked, the two predictions $f_{g}(a|b)$ and $f_{g}(a)$ should have a larger difference as the features of node $b$ will influence the prediction of node $a$ through the link $(u,v)$.
Formally, if $||f_{g}(a|b)-f_{g}(a)||_2>\eta$, where $\eta$ is an empirical threshold, the attacker determines that there is a link between $u$ and $v$. 

{While influence-based attacks have shown better performance than similarity-based attacks in certain scenarios~\cite{wu2022linkteller, meng2023devil}, their practical applicability remains limited in certain contexts. 
First, these attacks are not robust to graph dynamics, as the influence is primarily determined by a fixed threshold rather than the attributes of the injected node. 
For instance, in social network scenarios, consider two users, $u$ and $v$, who are not linked. The attacker follows the method in~\cite{meng2023devil}, linking malicious users $a$ and $b$ to target users $u$ and $v$. 
The attack observes the influence of $b$ on $a$ by calculating the difference between the predictions $f_{g}(a|b)$ and $f_{g}(a)$. 
However, if a normal user links to $u$ (follows $u$), a common occurrence in real-world applications, it can cause a significant difference between $f_{g}(a|b)$ and $f_{g}(a)$. 
This misleads the attacker into identifying the change as the influence of $b$ on $a$ through the non-existent link between $u$ and $v$, resulting in substantial false predictions under practical conditions.
Second, these attacks rely on querying the GNN model with newly injected nodes that have not been part of the training process, rendering them unsuitable for transductive models.
Given these considerations, our work focuses primarily on defending GNN models from similarity-based link-stealing attacks.
However, we will also evaluate the effectiveness of our proposed defense against influence-based attacks in Section~\ref{subsec:exp_influence}.}

\thispagestyle{empty}
\vspace{-0.5em}
\section{Problem Statement}
\vspace{-0.5em}
\label{sec:Problem} 

This section clarifies our threat model and defense objectives and then formulates our defense problem. 

\vspace{-0.5em}
\subsection{Threat model}
\vspace{-0.5em}

\smallskip
\noindent{\bf Target Model.}
We focus on the GNN model $f_g()$ that has been trained on a graph dataset $\mathcal{G}=(\mathcal{N}, \mathcal{A}^{\mathcal{N}},\mathcal{F},\mathcal{L})$, for node classification tasks.
Given a node $n_i$ as a querying sample, the trained GNN model will output a prediction vector $v_i$, representing the probability distribution that the querying node belongs to each class.

\smallskip
\noindent{\bf Attacker.}
The attacker's goal is to perform similarity-based link-stealing attacks, attempting to determine whether there exists a link between two nodes. 
The attacker has black-box access to the target GNN model $f_g()$. 
He can query the model with nodes to obtain their prediction vectors, which are then leveraged to infer whether the two nodes are linked or not. 
We follow the same setting as in~\cite{he2021stealing}, by considering three types of background knowledge that an attacker may possess.
In addition, we consider the challenging scenario in which the attacker can know the defense mechanism adopted by the model provider to adapt his attack method.

\smallskip
\noindent{\bf Model Provider.}
The model provider trains the target GNN model $f_g()$ and makes it available for public or commercial query access. 
Under our threat model, the model provider's goal is to defend against potential link-stealing attacks in order to protect its training graph. 
Naturally, the model provider has complete knowledge of the target model $f_g()$ and its training graph. 
{On the other hand, regarding 
different link-stealing attacks,  
the model provider may not know exactly which type of attack is to be launched.}
Hence, to mitigate all potential threats, he has to develop a mechanism capable of defending against all types of attacks.

\vspace{-0.5em}
\subsection{Defense Formulation}
\vspace{-0.5em}
In this paper, we aim to propose a defense method for the model provider to protect its GNN model from link-stealing attacks. 
Our goal is twofold: 1) The defense method should prevent the potential link-stealing attack from inferring the presence or absence of a link between two querying nodes and 2) the target model's prediction utility should not be compromised. 
Next, we detail our motivation and provide mathematical formulations for our defense methods. 

\subsubsection{Our Motivations}
Although multiple attack schemes differ in implementation details to encompass various attack scenarios, they all leverage the same pattern: the similarity of prediction vectors from adjacent nodes is greater than that of non-adjacent nodes. 
As such, our defense method should aim to distort this similarity. 
By doing so, attacks that rely on distinguishing the similarity of an adjacent node pair from that of a non-adjacent node pair will become ineffective.

However, this similarity pattern arises from the data aggregation from neighboring nodes, which constitutes the fundamental design integral to the GNN model’s efficacy.
Dissociating the correlation between prediction similarity and the attributes of node adjacency during the model training process will inevitably degrade model performance. 
Therefore, we propose adding noise to the prediction vectors for the nodes in the training graph after the model has been trained to disguise the high similarity between adjacent nodes' prediction vectors. 
That is, the model provider can first train the GNN model and then append the noise to the prediction vector output from the trained GNN model.
Hence, the attacker can only obtain the noisy predictions for the training nodes, which makes it hard to steal the link. 
Formally, the noisy prediction vector ${\bf r}$ is expressed as 
\begin{equation}
	{\bf r}={\bf v}+{\bf s}\ ,
\end{equation} 
where ${\bf s}$ represents the noise vector calculated by our method corresponding to a prediction vector  ${\bf v}$ for a training node from the target GNN model. 
Additionally, we will formulate this noise generation as an optimization problem to ensure that both defense goals and model utility are maintained.

\subsubsection{Defense Goal}
To defend against the similarity-based link-stealing attack, the noise calculated for the prediction vectors of two neighboring nodes must reduce their high similarity.
For any two neighboring nodes $i$ and $j$, we quantify the similarity of their prediction vectors $\mathbf{v}_i $ and $\mathbf{v}_j$, denoted as $sim(\mathbf{v}_i,\mathbf{v}_j)$, as follows 
\begin{equation}
\label{similarity}	sim(\mathbf{v}_i,\mathbf{v}_j)=corr(\mathbf{v}_i,\mathbf{v}_j)+cos(\mathbf{v}_i,\mathbf{v}_j)\ ,
\end{equation}
where $corr()$ and $cos()$ measure the Pearson correlation coefficient and cosine metrics, with detailed calculations shown in Appendix~\ref{sec:SimCal}.
We choose correlation and cosine metrics since they were observed in~\cite{he2021stealing} to better facilitate pattern extraction for link-stealing attacks.
Thus, our defense objective can be formulated as an optimization problem of noise calculation for similarity minimizing, yielding:
\begin{equation}
	\label{goal1}
	\text{OPT-1: min } \sum_{n_i.n_j\in\mathcal{N},\mathcal{A}_{ij}=1}sim(\mathbf{r}_i,\mathbf{r}_j)\ .
\end{equation}

\subsubsection{Utility Goal}
To ensure that the target model's prediction performance is not compromised, we introduce three constraints that consider the model's prediction utility from different perspectives, as detailed below.

\smallskip
\noindent{\bf Prediction Loss.}
We first consider the prediction loss, which measures the label predicted by a GNN model for the queried node. 
Preventing prediction loss is critical since, in some applications such as healthcare, even 1\% of wrongly predicted labels are intolerable and may incur irreparable consequences. 
Hence, to maintain the target model's prediction performance, we aim to achieve zero prediction loss in our defense mechanism, i.e., our added noises should not alter the predicted label for any queried node.
Recall that the predicted label class of a node is determined by the largest confidence score in its prediction vector. 
Such a prediction loss constraint can be expressed by 
\begin{equation}
\label{Loss1}
\arg\max_{a}\{v^a\}=\arg\max_{a}\{v^a+s^a\}\ ,
\end{equation}
where $v^a$ and $s^a$ respectively represent the $a$-th element in the prediction vector $\mathbf{v}$ and in the noise vector $\mathbf{s}$.
Eqn.~(\ref{Loss1}) constrains that the largest element in a prediction vector will remain the same before and after adding the noises, thus keeping the label unchanged.

\smallskip
\noindent{\bf Probability Distribution.}
The second utility constraint is that the prediction vector should remain a probability distribution, with each element representing the probability of being classified into a particular class, and the sum of all elements is equal to 1.
The newly added noise vectors should follow this property, yielding: 
\begin{equation}
		\label{Loss2}
	0\leq v^a+s^a\leq1, \sum_{a}(v^a+s^a)=1,\forall v^a\in\mathbf{v}, s^a\in\mathbf{s}\ .
\end{equation}

\smallskip
\noindent{\bf Distortion Budget.}
The third utility constraint aims to preserve latent information in the prediction vectors, which are deemed important when applying them as feature vectors for other downstream tasks. 
Hence, we need to minimize the difference between the original and altered prediction vectors. 
Specifically, the model provider can specify an allowable distortion budget $\theta$, representing the maximum alteration to any prediction vector for constraining altered latent information. 
Mathematically, we have 
\begin{equation}
		\label{Loss3}
	d(\mathbf{v},\mathbf{v}+\mathbf{s})\leq\theta\ ,
\end{equation}
where $d()$ is the distance calculation to measure the difference between the original and distorted prediction vectors, with the L-1 norm adopted for such distance calculation. 

\subsection{Defense  Formulation}
\vspace{-0.5em}
Based on our goals, we can mathematically formulate our defense scheme as an optimization problem, i.e.,  
\begin{align*}
	\text{OPT-2:\ }	\min\ &\sum_{n_i,n_j\in\mathcal{N},\mathcal{A}_{ij}=1}sim(\mathbf{v}_i,\mathbf{v}_j)\notag\\
	s.t.\ &(\ref{similarity}), (\ref{Loss1}), (\ref{Loss2}), \text{and }(\ref{Loss3}) ,\notag\\
\end{align*}
where prediction vectors $v$ in the formulation are constant and are produced by the trained GNN model. 
The noise vectors $s$  are variables to be solved.
This formulation aims to find the optimal noise vectors to achieve the defense goal.

\section{GRID: Graph Link Disguise}
\label{sec:GRID}
This section delves into the challenges in solving the proposed optimization problem and presents our intuitions for resolving them, i.e., our Graph Link Disguise (GRID).

\vspace{-0.5em}
\subsection{Key Challenges}
\vspace{-0.5em}
\label{sec:challenges}
Three challenges arise when solving the optimization problem OPT-2 in practice.

The first challenge arises from the inherent difficulty within GNNs of dissociating the correlation between prediction similarity and the attributes of adjacent nodes.
The message passing and data aggregation mechanisms of the GNN model inevitably cause an apparently high degree of similarity between the prediction vectors of adjacent nodes. 
For example, our experiments on the Cora dataset \cite{welling2016semi} reveal that the mean cosine similarity is 0.94 for adjacent nodes, compared to 0.63 for non-adjacent nodes. 
While adding heavy noise vectors to the prediction vectors can reduce this similarity, as formulated by our mathematical constraints, only slight noise can be tolerated to preserve the model's utility and prediction performance. 
Thus, the significant difference in similarity due to the GNN poses considerable challenges in distorting the similarity among adjacent nodes while maintaining utility.

The second challenge originates from the computational complexity and noise offset arising from the incorporation of noise. 
In OPT-2, all nodes are considered for noise addition. 
If we simultaneously optimize the noise vectors for all nodes, the computational complexity and time cost become unacceptable in practice, especially with a large number of nodes in the training graph. 
On the other hand, if we calculate the noise individually for each pair of neighboring nodes, the independently crafted noise vectors may counteract each other during similarity calculations, rendering them ineffective in preventing an attacker from inferring links. 
This issue is exacerbated when a node is linked to multiple nodes in the graph. 
Therefore, when crafting noise vectors, it is crucial to consider the graph topology to address these problems, posing a significant challenge in our design.

The third challenge stems from the need to take into account the potential adaptive attacks. 
We consider the most challenging scenario, where the attacker has complete knowledge of our defense scheme, thereby being able to conduct a tailored adaptive attack. 
If we blindly minimize the similarity between adjacent nodes, the attacker can recognize such a strategy and infer the presence of a link when the similarity of the prediction vectors of a querying node pair is below a certain level. 
Hence, enabling our defense to counter the adaptive attack is challenging.

\subsection{Our Intuitions}
\vspace{-0.5em}
 
To tackle the first and third challenges, we propose the Graph Link Disguise solution which focuses on diminishing the prediction similarity values of adjacent nodes to be commensurate with those of n-hop indirect neighboring nodes rather than minimizing them as much as possible.
This commensurate way is more attainable as the required reduction in the similarity of adjacent nodes is less, able to effectively cope with the first challenge.  
Besides, with the prediction similarity of adjacent nodes akin to that of n-hop indirect neighboring nodes rather than becoming lowest, this method can camouflage adjacent nodes as indirectly connected neighboring nodes, able to defend against adaptive attacks as described in the third challenge. 
Hence, our goal becomes to minimize the gap between the similarity of adjacent nodes and that of n-hop indirect neighboring nodes.

To tackle the second challenge incurred by noise addition on all nodes, we select a subset of nodes to form a core node set that encompasses nodes linked to all graph nodes. 
This set is known as the vertex cover set in graph theory \cite{Vertex}, which includes at least one endpoint node of every edge in the graph. 
Since all nodes are linked to these core nodes, the similarity of every pair of adjacent nodes can be altered by the noise crafted for the core nodes' prediction vectors. 
By focusing only on these core nodes for noise crafting, we can substantially reduce computational complexity. 
Additionally, because only one node in each edge is considered, we can calculate the noise individually for each pair of neighboring nodes while mitigating noise offset.

\subsection{GRID Solution}
Following our intuitions, we rectify the optimization problem of OPT-2 in calculating the noise vectors.

\subsubsection{Link Disguise}

Here, we achieve the link disguise by minimizing the gap between the similarity of adjacent nodes and of n-hop indirect neighboring nodes.
Considering a node $i$, its adjacent node set is denoted as $\mathcal{P}_{i}$,  and its n-hop indirect neighboring nodes are denoted as $\mathcal{Q}_{i}$.
The similarity gap denoted as $D_{i}$, under corresponding noise vector $\mathbf{s}_i$, is calculated by
\begin{equation}
	\label{distortion2}
	D_{i}=\sum_{j\in\mathcal{P}_i}sim(\mathbf{v}_i+\mathbf{s}_i,\mathbf{v}_j)-\sum_{k\in\mathcal{Q}_i}sim(\mathbf{v}_i+\mathbf{s}_i,\mathbf{v}_k)\ .
\end{equation}
This formulation captures the gap between the similarity of a node $i$ to its adjacent nodes and the similarity of this node to its n-hop indirect neighboring nodes, achieved by adding the noise vector $s_i$.
As such, the objective function in OPT-2 can be transformed into the following form:
\begin{equation}
	\min \sum_{i\in\mathcal{N}}D_{i}\ .
\end{equation}
This objective function aims at minimizing the similarity gap corresponding to all nodes.

\subsubsection{Threshold-based Core Nodes Selection}
\label{subsec:Core}

We denote the set of core nodes as $\mathcal{N}_c$, which should cover all links in the training graph $\mathcal{G}$. 
That is, any link in the graph should connect to at least one node in $\mathcal{N}_c$. 
Intuitively, we should identify the minimum vertex cover set as the core nodes so that the number of noise vectors added by GRID is minimized.
However, finding the minimum vertex cover set is an NP-hard problem.
Fortunately, our scenario does not require the core node set to be exactly the same as the minimal vertex cover set, due to the following reasons. 
First, our objective is to degrade the similarity level of adjacent nodes to that of n-hop indirect neighboring nodes. 
If the similarity of a given linked node pair is already low enough (e.g., lower than the average similarity of n-hop indirect neighboring nodes), it is unnecessary to include its corresponding nodes. 
Second, a high-degree node is linked with more nodes, resulting in a broad range of similarity values for its related links.
Selecting these high-degree nodes would require substantial amounts of noise (referred to as heavy noise) to modify their prediction vectors and alter the similarity values of all connected nodes. 
Third, from a mathematical perspective, the cosine similarity metric we employ is highly sensitive to the original value, requiring more effort (i.e., heavy noise) to degrade smaller cosine similarity values.
Considering these factors, we propose a threshold-based core node selection approach, by setting a threshold $\delta$ as the averaged similarity value of n-hop indirect neighboring nodes. 

\begin{algorithm}
	\caption{Core Node Selection}
	\LinesNumbered 
	\KwIn{$\mathcal{G}$, $\mathcal{C}$, $\delta$}
	\KwOut{$\mathcal{N}_c$}
	Extract the edge $\mathcal{E}$ from the adjacency matrix $\mathcal{A}$.\\
	\ForEach{$e_i\in\mathcal{E}$}{
		$e_i.weight=sim(\mathbf{v}_i,\mathbf{v}_j)$;
	}
        \ForEach{$n_i\in\mathcal{N}$}{
		$n_i.degree=\sum_{j\in\mathcal{N}}\mathcal{A}_{ij}\cdot sim(\mathbf{v}_i,\mathbf{v}_j)$;
	}
	Sort($\mathcal{E}$)\\
	\ForEach{$e_i\in\mathcal{E}$}{
		\If{$e_i.weight<\delta$}{
			countine;\\
		}
		\If{$n_i, n_j \notin \mathcal{N}_c$}{
			\If{$n_i.degree>n_j.degree$}{
				$\mathcal{N}_c = \mathcal{N}_c\cup n_i$
			}
			\Else{	$\mathcal{N}_c = \mathcal{N}_c\cup n_j$}
		}
	}
	\label{alg:core}	
\end{algorithm}

The proposed threshold-based core node selection approach is depicted next, with its pseudo-code listed in Algorithm~1.
The input is training graph $\mathcal{G}$ with the node set $\mathcal{N}$, adjacency matrix $\mathcal{A}$, and a preset threshold $\delta$.
Its output is a set of core nodes $\mathcal{N}_c$. 
{Specifically, the threshold $\delta$, determined by the averaged similarity value of n-hop indirect neighboring nodes, is approximated using a 
subset of n-hop indirect node samples from the training graph to reduce computational complexity.}
Our algorithm consists of three steps.
First, we extract the edge list $\mathcal{E}$ from the adjacency matrix $\mathcal{A}$ and calculate the similarity value of the neighboring nodes of each edge as the weight for sorting the edges in descending order.
Second, we sum the similarities of each node with all its adjacent nodes as its degree.
Third, we iterate through each edge in our sorted list to select the core node set.
That is, an edge in this list is dropped, if its similarity metric is less than the threshold $\delta$; otherwise, if both vertices of this edge are not yet in the vertex cover, we add the one with a higher degree to the core node set.
This process continues until all edges in the sorted list have been checked. 
The result is a set of core nodes $\mathcal{N}_c$ which cover all graph edges having weights above the defined threshold.
{The computation cost for Algorithm 1 is dominated by extracting the edge list and then calculating prediction similarity on each link.
    So, Algorithm~1 requires examining each node pair when the graph topology is stored in the adjacency matrix, which leads to a time complexity of $O(n^2)$ where $n$ is the number of nodes.
    The calculation of similarity has a complexity of $O(e)$ for all edges, where $e$ indicates the number of edges in the edge list.
    Hence, the total complexity is $O(n^2+e)$.}

\subsubsection{Reformulating Constraints}
We then reformulate constraints in the OPT-2 problem into solvable forms.

\smallskip
\noindent{\bf Prediction Loss.}
The prediction loss constraint (\ref{Loss1}) requires that the largest element in the prediction vector be unaltered after adding the noise vector. 
Hence, it can be rewritten as a set of 
 inequalities: $v^*+s^*-(v^a+s^a)\geq 0, \forall v^a\in\bf v$,
where $v^a$ and $s^a$ respectively represent entries in the prediction vector and in the noise vector, while  $v^*$ and $s^*$ refers to the largest element in the prediction vector $\bf v$ and the corresponding entry in the noise vector $s$.
We require the element $v^*+s^*$ to remain larger than all other entries. 

\smallskip
\noindent{\bf Probability Distribution.}
The prediction loss constraints ensure that the noisy vector follows a normal distribution. 
The first requirement, $0\leq v^a+s^a\leq1$, can be directly treated as an inequality constraint. 
The second requirement, $\sum_{a}(v^a+s^a)=1$, can be simplified as $\sum_{a}(s^a)=0$, since $\sum_{a}(v^a)=1$.

\smallskip
\noindent{\bf Distortion Budget.}
The distortion budget constraint~(\ref{Loss3})  controls the difference between the original prediction vector and the distorted prediction vector.
In this paper, we adopt the L-1 norm to measure the difference between two vectors, so this constraint can be rewritten as $	\| \mathbf{s}\|-\theta\leq 0$,
which represents that the sum of the absolute value of each element in the noise vectors should be less than the distortion budget.

\begin{algorithm}[t]
\SetAlgoLined
\caption{Iterative Gradient Descent for OPT-GRID with KKT Conditions}
\KwIn{Initial noise vector $s^{(0)}$, learning rate $\alpha$, tolerance $\epsilon$, maximum iterations $K$, {step size for Lagrange multipliers $\beta$} }
\KwOut{Optimized noise vector $s$}

Initialize $s^{(0)}$, $\lambda^{(0)}$, $\mu^{(0)}$, $\nu^{(0)}$\;
\For{$k = 0$ \KwTo $K$}{
   \text{Calculate }$\nabla_s \mathcal{L}$\;   
    $s^{(k+1)} = s^{(k)} - \alpha \nabla_s \mathcal{L}$\; 
    $s^{(k+1)} = \text{ConstraintCheck}(s^{(k+1)}, v_i, \theta)$\;
    Update $\lambda^{(k+1)}$, $\mu^{(k+1)}$, $\nu^{(k+1)}$ using:
    \begin{align*}
    \lambda^{(k+1)} &= \max\big(0, \lambda^{(k)} + \beta \left( v^a + s^{a} - v^* - s^{*} \right) \big) \\
    \nu^{(k+1)} &= \max\big(0, \nu^{(k)} + \beta \left( \| \mathbf{s} \| - \theta \right) \big) \\
    \mu^{(k+1)} &= \mu^{(k)} + \beta \left( \sum_a s^{a} \right)
    \end{align*}
    \If{$\| s^{(k+1)} - s^{(k)} \| < \epsilon$ }{
        \textbf{break}
    }
}
\Return $s = s^{(k+1)}$\;
\label{alg:KKT}
\end{algorithm}

\begin{algorithm}
\SetAlgoLined
\caption{ConstraintCheck Handling}
\KwIn{Noise vector $s$, Prediction vector $v_i$, Norm threshold $\theta$, and Vector dimension $A$}
\KwOut{Constrained noise vector $s$}

\textbf{Step 1: Handle Distribution Constraint L2:}

\[
s = s - \dfrac{1}{A} \left( \sum_{a=1}^{A} s^a \right) \mathbf{1}
\]
\For{$a = 1$ \KwTo $A$}{
    \[
    s^a = \min\left( \max\left( s^a, -v_i^a \right), 1 - v_i^a \right)
    \]
}

\textbf{Step 2: Handle Norm Constraint L3:}

\If{$\|\mathbf{s}\| > \theta$}{
    \[
    \mathbf{s} = \mathbf{s} \times \frac{\theta}{\|\mathbf{s}\|}
    \]
}

\textbf{Step 3: Handle Label Constraint L1:}

Let $C \gets \arg\max_{a} v_i^a$\;
\For{$a = 1$ \KwTo $A$}{
    \If{$a \neq C$ \textbf{and} $v_i^C + s^C - (v_i^a + s^a) < 0$}{
        $\delta \gets \frac{v_i^a + s^a - v_i^C - s^C}{2}$\;
        $s^a \gets s^a - \delta$\;
        $s^C \gets s^C + \delta$\;
    }
}
\Return $s$\;
\label{Alg:KKT2}
\end{algorithm}

\subsubsection{Reformulation of OPT-GRID}

Finally, we summarize all efforts to reformulate OPT-2 into the OPT-GRID.
{Given the prediction vectors of all nodes in the graph, for each node in the core node set, $n_i\in \mathcal{N}_c$, we calculate its noise vectors $s$ by solving the following optimization problem.
\begin{align*}
	\text{OPT-GRID:\ }\min_{s}\ & D_{i}\notag\\
	s.t.\  &\text{L1: } (v^a+s^a)-(v^*+s^*)\leq 0, \forall v^a\in\bf v_i\ ;\notag\\
	&\text{L2: }\sum_{a}(s^a)=0\ ;\notag\\
	&\ \ \ \ \ \ \ 0\leq v^a+s^a\leq1\ ;\notag\\
	&\text{L3: }\| \mathbf{s}\|-\theta\leq 0\ .\notag\
\end{align*}}
In this optimization problem, the optimized variable is the noise vector $s$, which impacts the calculation of the similarity gap $D_{i}$ according to Eqn.~(\ref{distortion2}).
{$s^*$ represents the entry in the noise vector ${\bf s}$ where the corresponding $v^*$ is the largest element in the prediction vector $\bf v_i$.}
OPT-GRID is a constrained non-linear optimization problem, known to be difficult to solve for a global optimal solution in polynomial time, especially considering the training graph with a large size and complex topology.
To address this problem, we apply the method of Lagrange multipliers, transforming the original constrained problem into an unconstrained one with a Lagrange function, as follows:
\begin{equation*}
    \mathcal{L}=D_{i}+\lambda(v^a+s^a-v^*-s^*)+\mu\sum_{a}(s^a)+\nu(\| \mathbf{s}\|-\theta)\ .
\end{equation*}
{Note that, the constraint $0\leq v^a+s^a\leq1$ is not appended with the Lagrange multiplier because it represents simple bound constraints that can be efficiently enforced through projection to improve computational efficiency.
Including this constraint in the Lagrangian would require introducing additional Lagrange multipliers for each bound, which is unnecessary and would complicate the optimization process without providing significant benefits.}
{Moreover, we use a single Lagrangian multiplier $\lambda$ for the L1 constraint to simplify the formula expression and enhance readability. 
In practice, a unique Lagrangian multiplier $\lambda^a$ is assigned for each $v^a\in\mathbf{v}_i$, ensuring that the L1 condition is satisfied for every $v^a$.}
We then employ the Karush–Kuhn–Tucker (KKT) conditions to regulate the optimization process. 
The KKT conditions include primal feasibility (the original constraints) with dual feasibility, complementary slackness, and stationarity, as shown below:
\begin{align*}
    &\lambda\geq 0 , \ \nu\geq 0;\\
    &\lambda(v^a+s^a-v^*-s^*)=0,\ \nu(\| \mathbf{s}\|-\theta)=0;\\
    &\nabla_s \mathcal{L}=0\ .
\end{align*}
{The calculation of $\nabla_s \mathcal{L}$ can refer to Appendix~\ref{sec:app:grad}.}
For each node in the core node set, we adopt the gradient descent method to iteratively update the noise vector, as outlined in Algorithm~\ref{alg:KKT}.
{The Lagrange multipliers are updated using the subgradient method to satisfy the dual feasibility and complementary slackness conditions of the KKT conditions and we control the adjustment magnitude to maintain stability and convergence.
More specifically, for $\lambda$ and $\nu$, the update increases their value when the constraint is violated and vice versa.
By updating them based on the constraint violations, the algorithm moves toward satisfying the complementary slackness conditions.
For $\mu$, the update adjusts its value based on the sum of $\sum_a s^{a}$, to ensure that $\sum_a s^{a}$ can move toward zero over iterations.}
Besides, the function $\text{ConstraintCheck}(s^{(k+1)}, v_i, \theta)$ ensures that the primal feasibility is satisfied by adjusting the solution $s^{(k+1)}$,  with the details illustrated in Algorithm~\ref{Alg:KKT2}.
{Especially, the constraint $0\leq v^a+s^a\leq1$ is enforced through the projection $s^a = \min\left( \max\left( s^a, -v_i^a \right), 1 - v_i^a \right)$.
In this adjustment, $\max\left( s^a, -v_i^a \right)$ ensures that $s^a$ is always no less than $-v_i^a$, i.e., $0\leq s^a+v_i^a$
Besides, $\min\left(s^a, 1 - v_i^a \right)$  ensures that $s^a$ is always no greater than $1 - v_i^a$, i.e.,  $v_i^a+s^a\leq1$.}

Once the update process converges, as indicated by changes in the noise vector smaller than $\eta$, or reaching the maximum iterations $K$, the algorithm terminates.  
{Following this process, we can find a near-optimal solution for the noise vector of each node in the core node set and then append them to the prediction vectors of the target model.}

\section{Performance Evaluation}
\label{sec:experiment}

We conduct experiments to assess GRID's performance in defending different types of link-stealing attacks across various datasets. 
Second, we compare GRID to its counterparts in terms of defending against link-stealing attacks. 

\subsection{Experiment Settings}
\label{sec:settings}

\subsubsection{Datasets}
We evaluate our approach on five widely adopted graph datasets: Citeseer~\cite{welling2016semi}, Cora~\cite{welling2016semi}, Pubmed~\cite{welling2016semi}, AIDS~\cite{caelli2002structural}, and ENZYMES~\cite{dobson2003distinguishing}, which are commonly used for assessing GNN models.


\subsubsection{Link Stealing Attacks}
\label{subsec:exp-attack}

{Our evaluations primarily focus on the first link-stealing attacks~\cite{he2021stealing}, as they provide the general and comprehensive framework for covering various attack scenarios. 
Among the eight proposed attacks, we highlight the performance of three representative attacks that encompass two key perspectives: unsupervised and supervised learning, as well as black-box and gray-box settings.
For the unsupervised Attack-0, we employ AUC (area under the ROC curve) as the primary evaluation metric to evaluate their performance. 
Additionally, to demonstrate the attack accuracy, recall, and precision metrics more concretely, we adopt the K-means clustering approach as implemented in~\cite{he2021stealing}. 
We set the value of K to 2, where the cluster with a lower (higher) averaged distance value is considered as the set of adjacent (non-adjacent) node pairs.
For the supervised black-box Attack-1 and supervised gray-box Attack-6, we follow their original settings 
and utilize their source codes~\cite{Stealcode} to implement the attack and calculate the AUC, accuracy, recall, and precision with and without incorporating our GRID defense mechanism.
Additionally, in Section~\ref{subsec:ablation} and~\ref{sub:GRID_settings},  we also include the other five attacks proposed in~\cite{he2021stealing} to ensure a comprehensive evaluation.
Second, we also consider the latest attack~\cite{wu2024link}, i.e., ILS, which is designed toward the inductive GNN model.
Specifically, we select attacks A2 and A9, which were reported as the strongest in the posterior-only and combined attack categories, to demonstrate our GRID performance.
}
The details about these attacks, including the setting of the shadow model and the link stealing classifiers, can be referred to~\cite{he2021stealing} and~\cite{wu2024link}.

\begin{table*}[]
	\centering
	\caption{Link stealing Attack-0 performance (\%) on five datasets with and without our GRID defense. Note: `a/b' are the values under GCN and GAT}
	\vspace{-0.8em}
 \resizebox{\linewidth}{!}{%
	\begin{tabular}{l|cc|cc|cc|cc|cc}
		\toprule
		& \multicolumn{2}{c|}{Citeseer}      & \multicolumn{2}{c|}{Cora}                & \multicolumn{2}{c|}{Pubmed} & \multicolumn{2}{c|}{AIDS} & \multicolumn{2}{c}{ENZYMES} \\
		& No Defense & GRID                 & No Defense & GRID                       & No Defense      & GRID     & No Defense     & GRID    & No Defense      & GRID      \\
		\midrule
		Attack Accuracy & 86.7 / 86.9 & 65.8 / 66.1 & 85.2 / 85.1 & 63.2 / 62.8 & 78.5 / 78.8 & 61.4 / 61.8 & 75.1 / 74.7 & 60.1 / 60.5 & 74.2 / 73.8 & 57.9 / 57.6 \\
		Attack Precision & 77.6 / 77.4 & 66.1 / 66.4 & 76.8 / 76.5 & 63.2 / 63.6 & 69.4 / 69.6 & 57.2 / 57.6 & 52.7 / 52.3 & 51.0 / 50.7 & 53.1 / 53.6 & 51.4 / 51.0 \\
		Attack Recall   & 98.1 / 97.7 & 64.3 / 64.7 & 95.7 / 96.1 & 63.1 / 63.3 & 94.5 / 94.2 & 63.2 / 62.8 & 97.6 / 97.3 & 64.5 / 64.8 & 98.7 / 98.4 & 63.3 / 63.1 \\
		Attack AUC      & 94.2 / 93.8 & 69.1 / 68.8 & 93.0 / 92.6 & 68.5 / 68.8 & 86.9 / 86.5 & 67.5 / 67.8 & 70.1 / 70.5 & 59.2 / 59.6 & 63.1 / 63.4 & 56.4 / 56.1 \\
        \midrule
		Model Accuracy & 73.3 / 75.8 & 73.3 / 75.8 & 84.8 / 85.6 & 84.8 / 85.6 & 81.6 / 83.3 & 81.6 / 83.3 & 68.5 / 70.5 & 68.5 / 70.5 & 69.0 / 71.4 &69.0 / 71.4\\
		\bottomrule    
	\end{tabular}
 }
	\vspace{-1em}
	\label{Tab:attack-0}
\end{table*}

\begin{table*}[]
	\centering
	\caption{Link stealing Attack-1 performance (\%) on five datasets with and without our GRID defense. Note: `a/b' are the values under GCN and GAT}
	\vspace{-0.8em}
 \resizebox{\linewidth}{!}{%
	\begin{tabular}{l|cc|cc|cc|cc|cc}
		\toprule
		& \multicolumn{2}{c|}{Citeseer}      & \multicolumn{2}{c|}{Cora}                & \multicolumn{2}{c|}{Pubmed} & \multicolumn{2}{c|}{AIDS} & \multicolumn{2}{c}{ENZYMES} \\
		& No Defense & GRID                 & No Defense & GRID                       & No Defense      & GRID     & No Defense     & GRID    & No Defense      & GRID      \\
		\midrule
		Attack Accuracy & 90.1 / 89.7 & 67.5 / 67.8 & 87.2 / 87.5 & 66.3 / 66.0 & 85.1 / 84.7 & 62.4 / 62.7 & 71.5 / 71.8 & 55.0 / 55.3 & 69.0 / 69.3 & 53.2 / 53.5 \\
		Attack Precision & 87.1 / 86.9 & 67.1 / 67.4 & 85.4 / 85.2 & 66.8 / 66.5 & 77.5 / 77.3 & 62.1 / 62.4 & 72.5 / 72.2 & 55.2 / 55.0 & 68.3 / 68.0 & 54.1 / 54.3 \\
		Attack Recall   & 95.8 / 95.5 & 68.4 / 68.2 & 88.3 / 88.0 & 65.8 / 66.1 & 89.1 / 89.4 & 62.6 / 62.4 & 70.1 / 70.3 & 54.7 / 54.5 & 70.1 / 70.3 & 52.7 / 53.0 \\
		Attack AUC      & 96.5 / 96.3 & 73.2 / 73.5 & 94.2 / 93.9 & 72.8 / 73.1 & 88.5 / 88.2 & 70.6 / 70.4 & 72.9 / 73.1 & 58.9 / 58.7 & 74.5 / 74.2 & 60.2 / 60.0 \\
        \midrule
		Model Accuracy & 73.3 / 75.8 & 73.3 / 75.8 & 84.8 / 85.6 & 84.8 / 85.6 & 81.6 / 83.3 & 81.6 / 83.3 & 68.5 / 70.5 & 68.5 / 70.5 & 69.0 / 71.4 &69.0 / 71.4\\
		\bottomrule    
	\end{tabular}
 }
	\vspace{-0.8em}
	\label{Tab:attack-1}
\end{table*}

\begin{table*}[]
	\centering
	\caption{Link stealing Attack-6 performance (\%) on five datasets with and without our GRID defense. Note: `a/b' are the values under GCN and GAT}
	\vspace{-0.8em}
  \resizebox{\linewidth}{!}{%
	\begin{tabular}{l|cc|cc|cc|cc|cc}
		\toprule
		& \multicolumn{2}{c|}{Citeseer}      & \multicolumn{2}{c|}{Cora}                & \multicolumn{2}{c|}{Pubmed} & \multicolumn{2}{c|}{AIDS} & \multicolumn{2}{c}{ENZYMES} \\
		& No Defense & GRID                 & No Defense & GRID                       & No Defense      & GRID     & No Defense     & GRID    & No Defense      & GRID      \\
		\midrule
		Attack Accuracy & 92.2 / 92.3 & 69.1 / 69.4 & 89.9 / 90.1 & 69.5 / 69.3 & 91.1 / 90.9 & 68.8 / 68.6 & 94.3 / 94.2 & 68.2 / 68.3 & 82.4 / 82.6 & 65.0 / 64.9 \\
		Attack Precision & 90.1 / 90.0 & 69.1 / 68.9 & 87.8 / 87.6 & 69.7 / 69.6 & 90.3 / 90.2 & 69.7 / 69.5 & 90.7 / 90.8 & 69.5 / 69.6 & 77.0 / 76.9 & 65.3 / 65.1 \\
		Attack Recall   & 93.3 / 93.5 & 69.2 / 69.0 & 93.0 / 93.2 & 69.3 / 69.5 & 92.4 / 92.3 & 68.5 / 68.7 & 98.6 / 98.5 & 67.5 / 67.6 & 88.7 / 88.5 & 64.8 / 64.9 \\
		Attack AUC      & 98.1 / 98.2 & 71.1 / 71.0 & 96.4 / 96.3 & 70.4 / 70.5 & 97.0 / 96.9 & 70.8 / 70.7 & 97.9 / 97.8 & 73.9 / 74.0 & 89.1 / 89.0 & 68.1 / 68.3 \\
        \midrule
		Model Accuracy & 73.3 / 75.8 & 73.3 / 75.8 & 84.8 / 85.6 & 84.8 / 85.6 & 81.6 / 83.3 & 81.6 / 83.3 & 68.5 / 70.5 & 68.5 / 70.5 & 69.0 / 71.4 &69.0 / 71.4\\
		\bottomrule    
	\end{tabular}
 }
	\vspace{-1em}
	\label{Tab:attack-6}
\end{table*}
\begin{table*}[]
	\centering
	\caption{Link stealing Attack-6 performance (\%) on five datasets with and without our GRID defense.}
	\vspace{-0.8em}
  \resizebox{0.9\linewidth}{!}{%
	\begin{tabular}{l|cc|cc|cc|cc|cc}
		\toprule
		& \multicolumn{2}{c|}{Citeseer}      & \multicolumn{2}{c|}{Cora}                & \multicolumn{2}{c|}{Pubmed} & \multicolumn{2}{c|}{AIDS} & \multicolumn{2}{c}{ENZYMES} \\
		& No Defense & GRID                 & No Defense & GRID                       & No Defense      & GRID     & No Defense     & GRID    & No Defense      & GRID      \\
		\midrule
		Attack Accuracy & 92.2 & 69.1 & 89.9 & 69.5 & 91.1 & 68.8 & 94.3 & 68.2 & 82.4 & 65.0 \\
		Attack Precision & 90.1 & 69.1 & 87.8 & 69.7 & 90.3 & 69.7 & 90.7 & 69.5 & 77.0 & 65.3 \\
		Attack Recall   & 93.3 & 69.2 & 93.0 & 69.3 & 92.4 & 68.5 & 98.6 & 67.5 & 88.7 & 64.8 \\
		Attack AUC      & 98.1 & 71.1 & 96.4 & 70.4 & 97.0 & 70.8 & 97.9 & 73.9 & 89.1 & 68.1 \\
        \midrule
		Model Accuracy & 73.3 & 73.3 & 84.8 & 84.8 & 81.6 & 81.6 & 68.5 & 68.5 & 69.0 & 69.0 \\
		\bottomrule    
	\end{tabular}
 }
	\vspace{-1em}
	\label{Tab:attack-6}
\end{table*}

\begin{table*}[htb]
	\centering
	\caption{The performance (\%) of two ILS attacks, A2 and A9, on GAT across five datasets, with and without our GRID. A2 and A9 are reported as the strongest in the posterior-only and combined attack categories, respectively}
	\resizebox{0.9\linewidth}{!}{%
	\begin{tabular}{cl|cc|cc|cc|cc|cc}
		\toprule
		&& \multicolumn{2}{c|}{Citeseer}      & \multicolumn{2}{c|}{Cora}                & \multicolumn{2}{c|}{Pubmed} & \multicolumn{2}{c|}{AIDS} & \multicolumn{2}{c}{ENZYMES} \\
		&& No Defense & GRID                 & No Defense & GRID                       & No Defense      & GRID     & No Defense     & GRID    & No Defense      & GRID      \\
		\midrule
		\multirow{4}{*}{ILS A2}& Accuracy & 82.54  & 66.21  & 83.71  & 67.41  & 82.71  & 64.27  & 75.44  & 60.68  & 75.51  & {60.01}  \\
            & Precision & 82.03  & 66.87  & 83.15  & 68.22  & 83.28  & 64.51  & 76.03  & {60.34}  & 74.79  & 60.35 \\
        & Recall   & 82.85  & 66.76  & 82.87  & 67.04  & 83.12  & 64.79  & 75.08  & 61.13  & 75.94  & {60.55}  \\
        & AUC      & 82.51  & 60.60  & 84.50  & {61.02}  & 80.92  & {57.01}  & 82.72  & {{57.81}}  & 81.24  & {57.90} 
    \\
    \midrule
        \multirow{4}{*}{ILS A9}& Accuracy & 92.45  & 67.53  & 91.76  & 68.84  & 90.67  & 66.58  & 84.47  & {63.58}  & 82.50  & 64.01  \\
		& Precision & 91.96  & 67.92  & 91.25  & 69.44  & 90.51  & 67.24  & 84.05  & {63.79}  & 83.00  & 64.59
  \\
		& Recall   & 92.72  & 66.98  & 91.85  & 68.31  & 91.14  & 65.07  & 84.80  & 63.20  & 82.21  & {63.06}
 \\
        & AUC      & 90.98  & 67.01  & 91.05  & 64.04  & 93.89  & 64.17  & 84.07  & 63.42  & 85.68  & {61.26} \\
  \midrule
	\multicolumn{2}{c|}{Model Accuracy} & {75.80} &  {75.80} & {85.60} &  {85.60} &  {83.30} &  {83.30} & {70.50} &  {70.50} &  {71.41} &{71.41}\\
		\bottomrule    
	\end{tabular}
 }
	\label{Tab:latest}
\end{table*}

\subsubsection{Dataset Configurations}
\label{subsec:dataset}

Regarding AIDS and ENZYMES, which are designed for graph classification, we assign the labels of the graph (molecular compounds) to nodes (molecules) as their ground truth labels.
During our experiments, we need to train a target model, implement the link stealing attacks, and evaluate the defense method, so we extract several disjointed sub-datasets from the original dataset D for experiments, depicted below. 

\smallskip
\noindent{\bf D1 (for training the target model).} 
We sample 40\% of the nodes and of the links from $D$ to compose the training graph $D1$.
{For the transductive training, e.g., GCN, we retained the full graph structure in D1 but removed 20\% nodes’ labels. 
The model was then trained to predict the labels of these unlabeled nodes. 
For the inductive training, e.g., GAT, we trained the model directly on D1 and let it predict unseen nodes in D.}

\smallskip
\noindent{\bf D2 (for performing attacks).} 
Some link-stealing attacks (i.e., Attack-1, Attack-4, Attack-5, and Attack-7) require a shadow dataset to perform the attack. 
We sample 40\% of the nodes and the corresponding links from $D$ to compose the shadow training graph $D2$.
It will be employed to train the shadow model for these attacks.
Note that not all the data samples in D2 are adopted to train the shadow model. 
The detailed adoption ratio can refer to~\cite{he2021stealing} and~\cite{Stealcode}.

\smallskip
\noindent{\bf D3 (for other defenses).} 
Existing defense methods may require the model provider to train a defender classifier to fit the attacker's inference classifier.
So, we sample 20\% of the nodes and the corresponding links from $D$ to compose the graph $D4$ for the defender classifier.

\smallskip
\noindent{\bf D4 (for testing attacks).} 
In D1, we sample $20\%$ node pairs that are linked and another $20\%$ node pairs that are unlinked to compose $D4$, for use to evaluate the link stealing attacks with and without defense.

\begin{table*}[h]
\centering
	\caption{Attack Accuracy and computational time under GRID with and without Core Node Selection. Note: `a/b' are the values under GCN and GAT}
 \resizebox{0.85\linewidth}{!}{%
\begin{tabular}{c|cccc|c}
\toprule
\multirow{2}{*}{} & \multicolumn{4}{c|}{w/ Core Node Selection} & \multirow{2}{*}{w/o Core Node Selection} \\
                  & n=2    & n=3    & n=4    & n=5       &                                          \\
                  \midrule
Attack Accuracy   &        &              &        &        &                                       \\
Attack-0          &  62.7 / 63.6      &  61.4 / 61.8      &   61.3 / 61.3     &  60.2 / 60.8          &    68.3 / 69.8                                       \\
Attack-1          &  63.4 / 64.2      &  62.4 / 62.7      &    61.8 / 61.6    &     60.4 / 59.7          &    68.4 / 68.8                                       \\
Attack-2          &   62.4 / 62.1     &  61.4 / 61.5      &    60.1 / 60.8    &   59.4 / 58.5            &   67.6 / 69.9                                       \\
Attack-3          &   64.0 / 63.9     &  63.3 / 63.3       &    62.0 / 61.9    &    60.2 / 60.1           &     68.4 / 68.7                                     \\
Attack-4          &  64.5 / 64.3      &  63.5 / 64.0      &   61.2 / 61.2     &   59.5 / 60.0            &      69.0 / 68.9                                    \\
Attack-5          &    67.9 / 66.8    &    65.9 / 66.2    &   63.5 / 64.0     &  62.0 / 61.8             &    70.1 / 71.9                                      \\
Attack-6          &     69.7 / 69.9     &  68.8 / 68.6       &   65.8 / 66.0       &         63.8 / 64.1        &     74.5 / 75.6                                     \\
Attack-7          &    65.8 / 66.2      &   64.9 / 65.3      &    62.9 / 63.0      &  60.9 / 61.8               &    70.7 / 71.4                                      \\
\midrule
Computation Time  &   6403.41s / 6391.24s     &  6823.70s / 6779.71s      &    7431.28s / 7558.92s    &    7937.75s / 7979.18s         &  45052.53s / 46148.31s                                        \\
\bottomrule
\end{tabular}%
}
\label{Tab:core_2}
\end{table*}

\subsubsection{Target Model}

We consider both transductive and inductive GNN models.
For the transductive one, we train the model of graph convolutional networks (GCN) for the node classification task, as described in~\cite{kipf2017semi}.
{Besides, for inductive models, we focused on GAT~\cite{velivckovic2018graph}, GraphSAGE~\cite{hamilton2017inductive}, and GIN~\cite{xu2018powerful}.}
We train the target model with D1 and sample another 20\% nodes in $D$ to evaluate model performance.
We implemented all these models based on publicly accessible code~\cite{GCN, GAT}.
The testing accuracy outcomes of target models on each dataset are similar to the results presented in the benchmark.
Our experiments are conducted on a workstation equipped with Intel Core i9-13900K and NVIDIA RTX 4090 GPU with 24GB of VRAM.

\subsection{Overall Performance on Different Attacks}

We employ our GRID to show defense performance results for different attack types under different models and datasets.
Here, we set the distortion budget $\theta=0.4$ and the hop $n=3$. 
It is important to note that in our context, a noise level of $\theta=0.4$ represents a relatively minor perturbation.
For example, $(-0.2, 0.2)$ is the largest noise for $\theta=0.4$ in the two-dimensional noise vector. 
Given that the prediction vectors for the target model exist in higher dimensions, such as 3 for Pubmed, 6 for Citeseer, and 7 for Cora, the resulting distortion on each confidence score remains quite minimal.  

{Regarding the first similarity-based attacks~\cite{he2021stealing}, the results for Attack-0, Attack-1, and Attack-6 on GCN and GAT are shown in Tables~\ref{Tab:attack-0},~\ref{Tab:attack-1}, and~\ref{Tab:attack-6}.
Additional results on GraphSAGE and GIN can be found in Appendix~\ref{subsec:other_results}.}
It is important to note that GRID does not degrade the target model accuracy, as shown in the last row of these tables. 
The prediction loss constraint in our optimization formulation ensures that the target model accuracy remains unchanged under GRID. 
Furthermore, our noise vector calculation does not require re-training the target model, so the training loss and expense remain unaffected. 

First, Table~\ref{Tab:attack-0} shows the performance of unsupervised Attack-0 under our GRID, where our GRID effectively defends against this attack across five datasets. 
The evaluation of unsupervised attacks heavily relies on the AUC metric, and a significant degradation of AUC scores was observed when our GRID was deployed. 
Notably, in the case of three citation datasets where attacks initially achieved superior performance, our GRID can incur an approximate 25\% degradation in their AUC metrics. 
Even for the two chemical datasets where attacks exhibited subpar performance, our GRID can reduce their AUC to around 60\%, rendering the corresponding attacks to random guessing.
Second, we turn our attention to the black-box attack, i.e., Attack-1, as depicted in Table~\ref{Tab:attack-1}, which showcases the performance variation under the influence of our GRID. 
The attack performance exhibits a significant degradation across all assessed metrics under the defense of our GRID.
Across the three citation datasets, we observed a degradation of approximately 20\% in all metrics, with a particularly noticeable deterioration of 25\% in the recall.
For the two chemical datasets, our GRID exhibits better performance,  reducing the attack performance to almost 50\%. 
Such an attack performance is only marginally better than random guessing. 
This significant reduction in attack performance underscores the effectiveness of our GRID in defending against black-box attacks.
Third, regarding the strongest gray-box attacks, Attack-6, our GRID performs a little bit inferior.
The reason is that these attacks can leverage the certain knowledge of node attributes and graph topology to launch strong attacks, while our GRID primarily focuses only on the prediction vectors.  
However, our GRID is still able to reduce its AUC performance to around 70\% with a distortion budget of only 0.4.

{Additionally, the results of A2 and A9 from the latest ILS attack~\cite{wu2024link} on GAT are shown in Table~\ref{Tab:latest}. 
To favor the attacks, the shadow models used to perform the attacks are identical to the target model. 
Despite this, we observe that our GRID method still significantly reduces the performance of various ILS attacks. 
This reason is that ILS also relies on similarity metrics calculated from the prediction vectors of pairwise nodes. 
By adding noise to the prediction vectors of the training nodes output with our GRID, the similarity patterns were distorted, thereby misleading the attack classifiers.}

\begin{figure}[t]
	\centering	
	\includegraphics[width=0.8\linewidth]{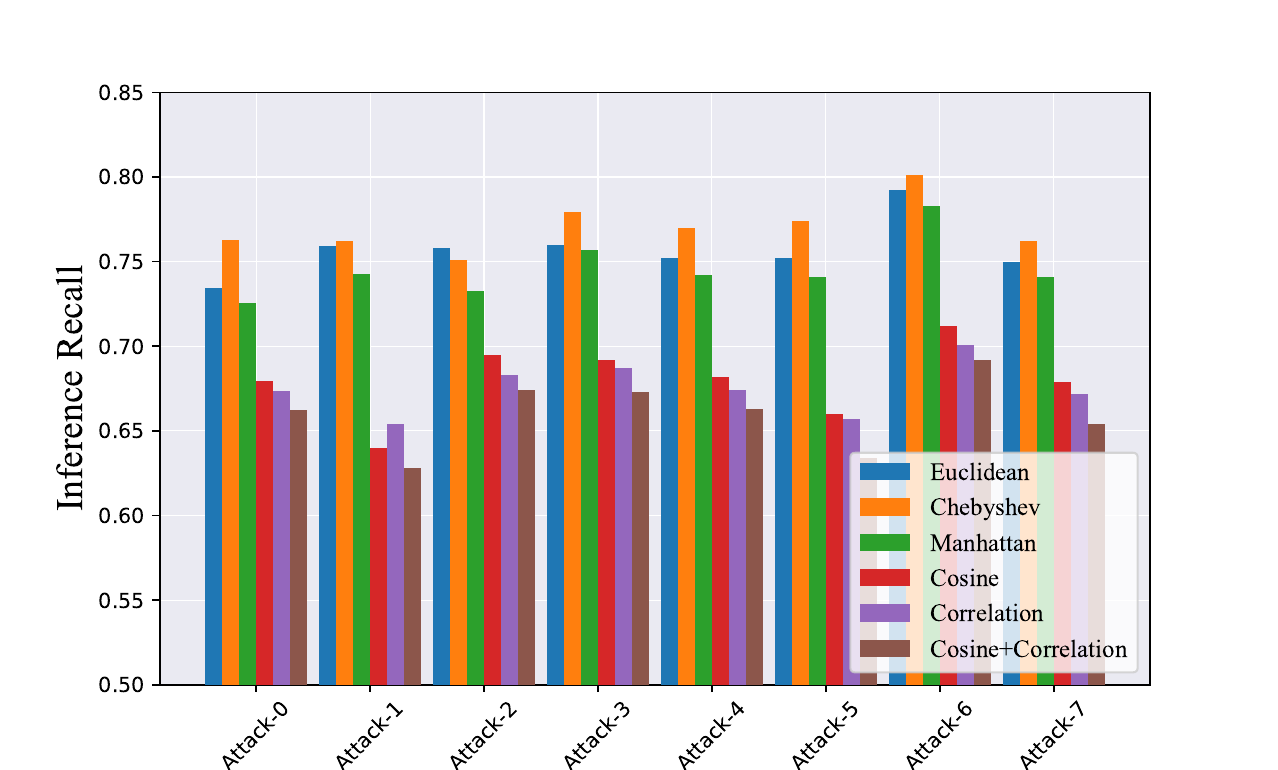}
	\vspace{-0.5em}	
	\caption{Link stealing attack recall variations for eight different attacks under different similarity metrics.}
	\vspace{-1.5em}
	\label{Fig:sim:citeseer}
\end{figure}

\subsection{Ablation Study}
\label{subsec:ablation}

\noindent{\bf Impact of Core Node Selection.}
The core node selection is a key design element in GRID that addresses computational complexity and noise offset issues. 
Therefore, we compare our GRID with and without this component to analyze its contribution to defense performance.
{For our GRID method with core node selection, we set $n$ between 2 and 5, and the noise is calculated only for the nodes in the core node set. 
Besides, the thresholds $\delta$, determined by the similarity of the prediction vectors of n-hop indirect neighboring nodes, are approximated using a maximum of 1,000 pairs of n-hop nodes.}
For GRID without core node selection, the noise is calculated for all nodes individually.
The distortion budget is set to $0.4$, and the maximum iteration for solving the optimization problem is set to 20. 
Here, we take Pubmed with 19717 nodes as an example, and 3201 nodes are chosen as the core node.

The results of computational time and the defense performance of eight attacks in~\cite{he2021stealing} are shown in Table~\ref{Tab:core_2}. 
First, we observe that the defense performance of GRID with core node selection is consistently better, as indicated by the lower attack accuracy across all eight attacks. 
This improvement is because, under GRID without core node selection, the noise for the two end nodes of each link tends to counteract during similarity disguise, as each noise is optimized for each node's neighboring nodes individually. 
{Second, the computational time for GRID with core node selection, which is less than 8,000 seconds, is significantly lower than that for GRID without core node selection, which exceeds 45,000 seconds. 
The computational time of GRID is highly dependent on the number of nodes selected for noise calculation. 
Under the core node selection scheme, only a subset of nodes needs to be calculated, compared to the total 19,717 nodes considered in GRID without core node selection. 
Additionally, since we sample a maximum of 1,000 pairs of n-hop nodes, the computational time for the similarity threshold calculation does not significantly increase as n increases. 
These two factors together result in substantial savings in computational time.
Furthermore, as $n$ increases, we observe a slight improvement in defense performance, although computation time also increases due to more nodes being selected as core nodes. 
This observation underscores the importance of selecting $n$ in practice to balance defense performance with computational time.
}

\noindent{\bf Impact of Different Similarity Metrics.}
Our current GRID design relies on the combination of correlation and cosine similarity measurement due to their superior performance in executing link-stealing attacks as reported in~\cite{he2021stealing}. 
Here, we adopt different similarity metrics and show their impacts on our GRID. 
In particular, the Euclidean, Chebyshev, Manhattan, Cosine, and Correlation are taken into account. 
That is, we replace our similarity calculation with each of the aforementioned metrics and evaluate the effectiveness of our GRID. 
Here, we set the distortion budget $\theta=0.3$, the $n=3$, and conduct experiments on the Citeseer dataset.
Our results are shown in Fig.~\ref{Fig:sim:citeseer}.

As observed in Fig.~\ref{Fig:sim:citeseer}, the combination of correlation and cosine consistently outperforms all other individual metrics across all attacks, which decreases the recall values at the maximum. 
For instance, in the case of Attack-6, the recall value for the combination of correlation and cosine is 0.692, while the recall values for Euclidean, Chebyshev, Manhattan, Cosine, and Correlation metrics are 0.7921, 0.801, 0.7831, 0.712, and 0.701, respectively.
On the other hand, we find that our defense performance under the Correlation or Cosine is always much better than that under other individual metrics. 
This further enhances our confidence to take the combination of correlation and cosine similarity rather than other individual or combined metrics. 

\begin{figure*}[h]
	\centering
	\subfigure[ Attack-0  ]{
		\includegraphics[width=0.23\linewidth]{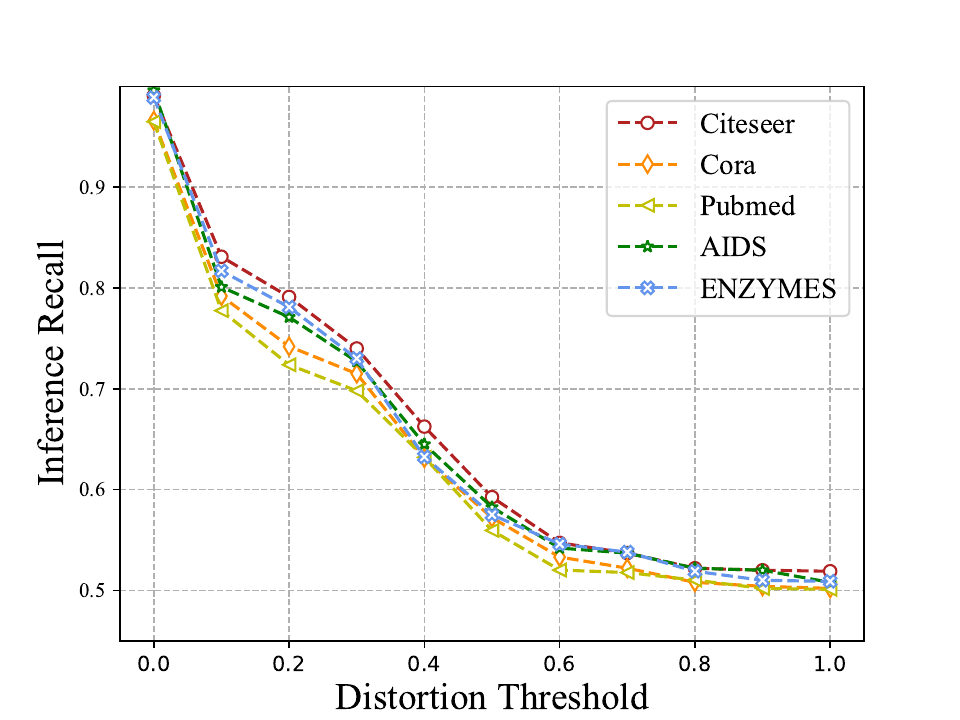}
		\label{Fig:attack0}
	}	\vspace{-0.5em}
	\subfigure[Attack-1]{
		\includegraphics[width=0.23\linewidth]{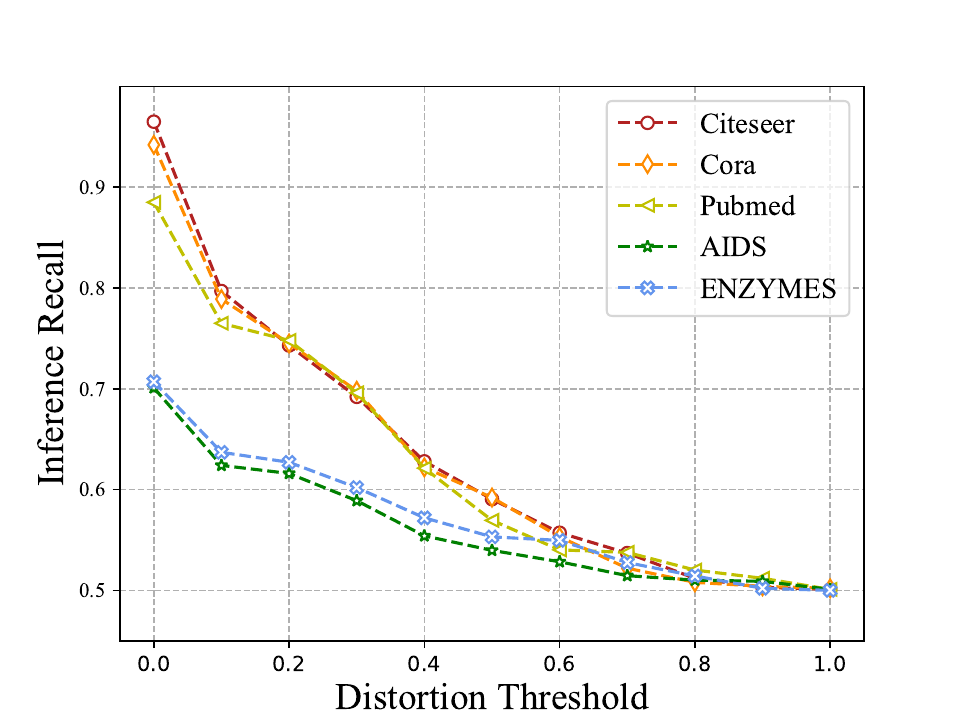}
		\label{Fig:attack1}
	}
	\subfigure[Attack-2]{
		\includegraphics[width=0.23\linewidth]{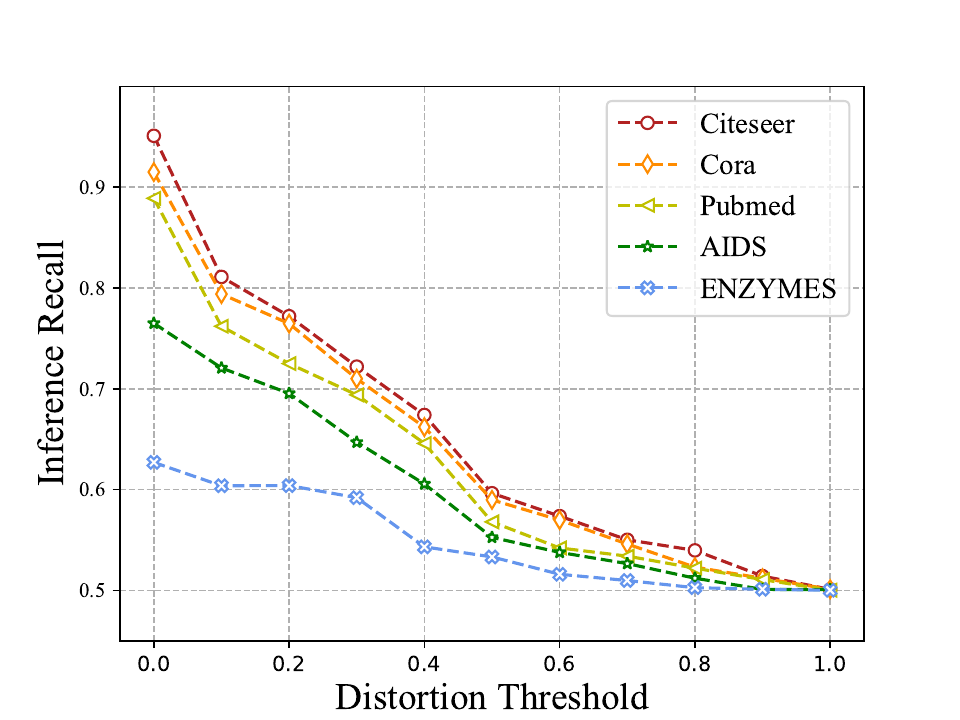}
		\label{Fig:attack2}
	}	
	\subfigure[Attack-3]{
		\includegraphics[width=0.23\linewidth]{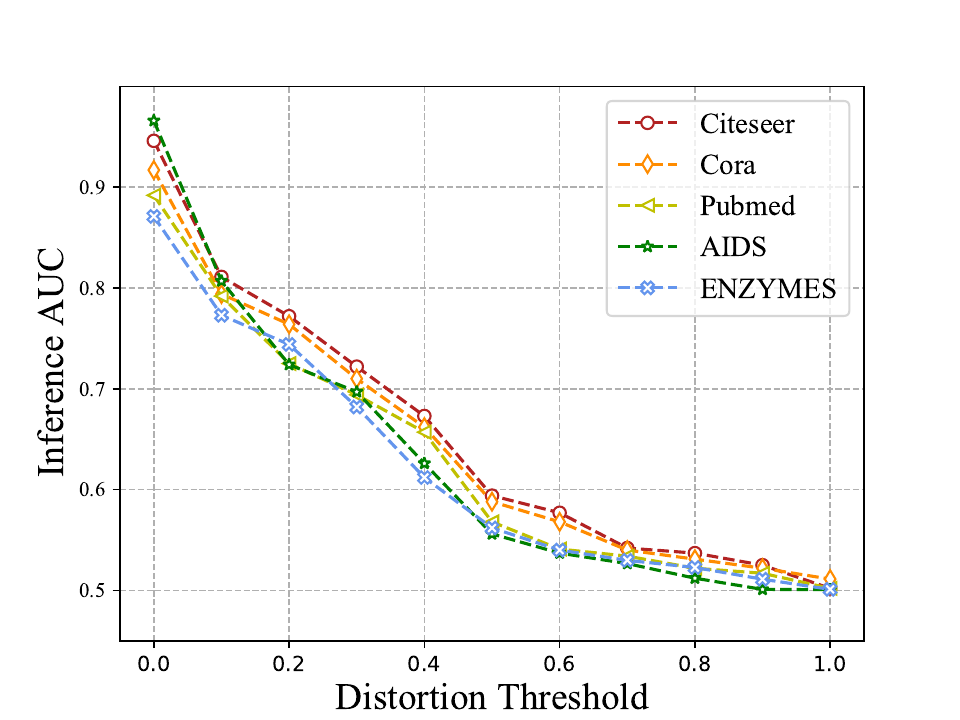}
		\label{Fig:attack3}
	}
	\subfigure[Attack-4]{
		\includegraphics[width=0.23\linewidth]{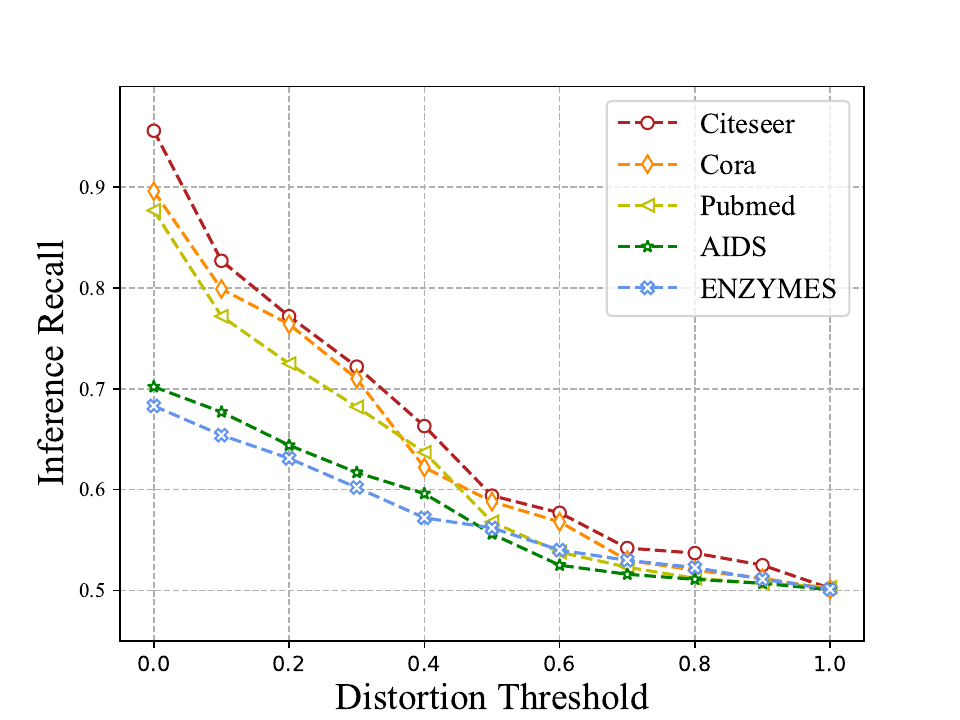}
		\label{Fig:attack4}
	}
	\subfigure[Attack-5]{
		\includegraphics[width=0.23\linewidth]{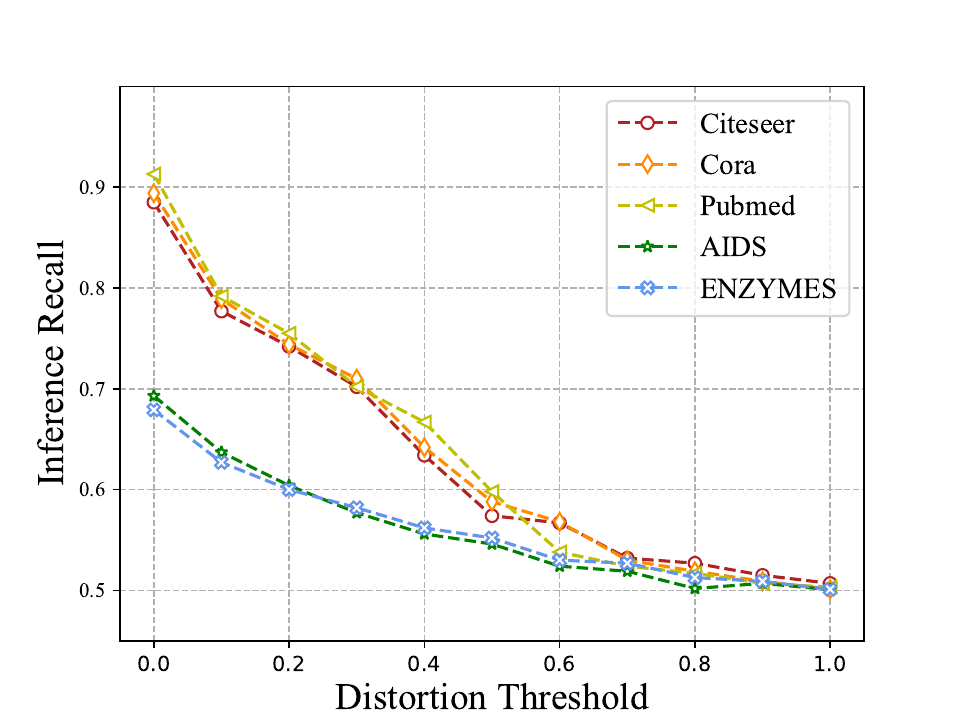}
		\label{Fig:attack5}
	}
	\subfigure[Attack-6]{
		\includegraphics[width=0.23\linewidth]{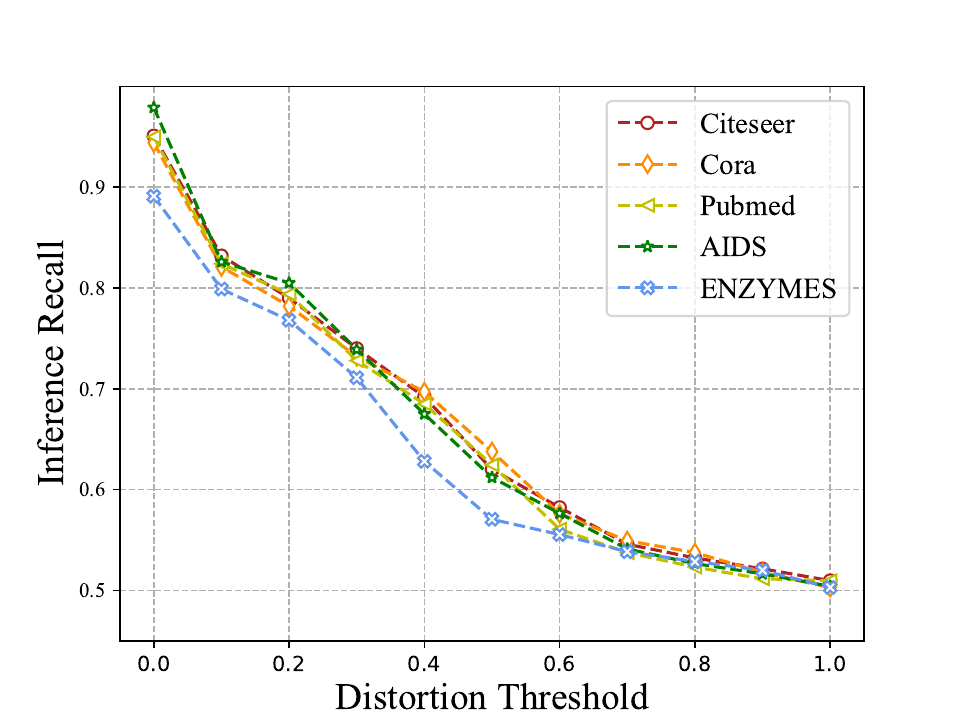}
		\label{Fig:attack6}
	}
	\subfigure[Attack-7]{
		\includegraphics[width=0.23\linewidth]{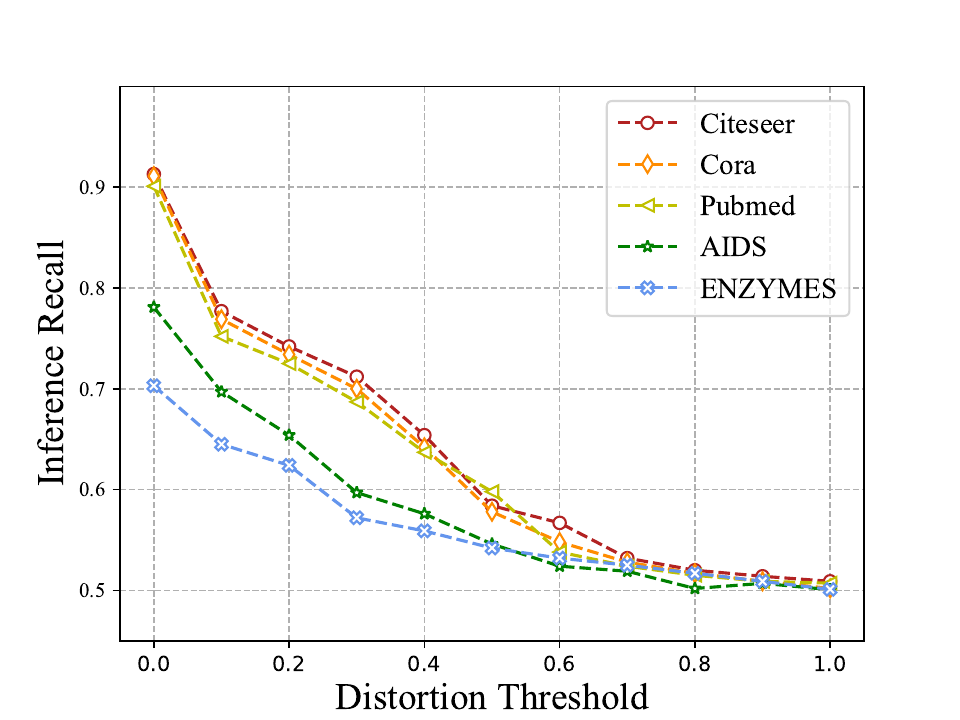}
		\label{Fig:attack7}
	}
	\vspace{-0.6em}
	\caption{Link stealing attack recall variations for eight different attacks under GRID with increasing distortion thresholds.}
	\label{fig:attack}
	\vspace{-1.5em}
\end{figure*}

\begin{figure}[h]
	\centering	
	\includegraphics[width=0.7\linewidth]{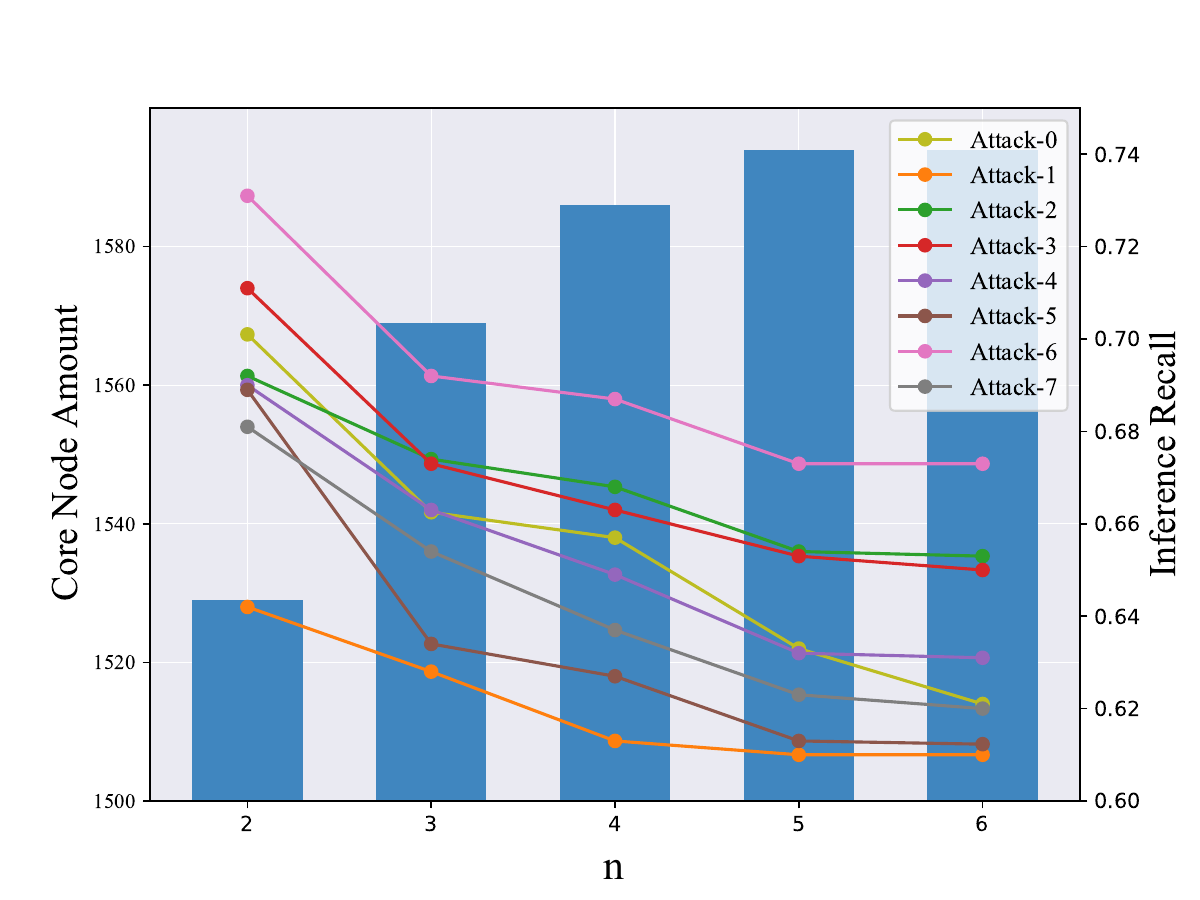}
	\vspace{-0.5em}	
	\caption{Link stealing attack recall and the core node set size variations for eight different attacks with increasing n.}
	\vspace{-1.5em}
	\label{Fig:N:citeseer}
\end{figure}

\subsection{GRID Performance under Different Settings}
\label{sub:GRID_settings}

Recall that there are two important control parameters for GRID: the distortion budget $\theta$, which controls the maximum distortion incurred by noise, and the value $n$, indicating the number of hops for indirect neighboring nodes, which influences both the similarity gap in the objective function and the threshold $\delta$ for core node selection.
Here, we analyze how these parameters impact our GRID.

\subsubsection{Defense Performance on Different Distortion Budget $\theta$}
The distortion budget $\theta$ represents the upper bound of the difference between the original prediction vector and the distorted one.
A larger budget signifies that the model provider can tolerate a larger distortion caused by the noise.
Here, we set the hop $n=3$ and vary $\theta$ from $0$ to $1$ with the step of $0.1$.
Since recall can measure the critical ability of whether an attacker can extract all links from the training graph, we will use recall as the primary metric to demonstrate our defense performance.
Figures~\ref{Fig:attack0} to~\ref{Fig:attack7} respectively show the recall variations regarding the attack performance of eight attacks on five datasets under the defense of our GRID when varying the distortion threshold $\theta$ on GCN.
We have the following observations.

\smallskip
\noindent{\bf \em Our GRID capability is enhanced with the increase in the distortion budget.}
From Figures~\ref{Fig:attack0} to~\ref{Fig:attack7}, we observe that the recall values of all eight link stealing attacks drop with an increase in the distortion budget.
In particular, all recall values on five datasets drop to around $50\%$ (similar to random guessing) when the distortion budget approaches 1.0. 
Taking the black-box Attack-1  as an example, i.e., Figure~\ref{fig:attack}(b), when $\theta$ increases from $0$ to $1$, the  recall values for Attack-1 drop from $96.5\%$ to $50.1\%$, from $94.2\%$ to $50.3\%$, from $88.5\%$ to $50.1\%$, from $70.1\%$ to $50.2\%$, and from $70.7\%$ to $50.2\%$, respectively, on five datasets.
Similarly, for the strongest attack, i.e., Attack-6,  the recall values of attack performance  drops from $95.1\%$ to $50.9\%$, from $84.4.8\%$ to $50.2\%$, from $85.0\%$ to $50.9\%$, from $97.9\%$ to $50.3\%$, and from $89.1\%$ to $50.4\%$, respectively, on five datasets, as shown in Figure~\ref{fig:attack}(g). 
The rationale behind this is that a larger distortion threshold allows more substantial noise vectors to be added to the prediction vectors, incurring a noticeable variation in the similarity of altered prediction vectors from two adjacent nodes. 
That is, the prediction similarity level between two adjacent nodes is akin to that of 3-hop indirect neighboring nodes, confusing link-stealing attacks.

\smallskip
\noindent{\bf \em GRID can incur quick decay for attack performance even with minor noise.}
We also observe that even with a small distortion budget i.e., $\theta=0.1$, the recall values still drop quickly for all attacks on all datasets. 
For example, the  recall values drop by $16.01\%$, $16.80\%$, $14.01\%$, $13.56\%$, $12.97\%$, $12.8\%$, $11.9\%$, and $13.6\%$ respectively on the Citesser dataset across all eight attacks.
This can deteriorate the attack performance from approximately $95\%$ to below $80\%$.
This observation demonstrates that our GRID can achieve notable defense performance even with negligible distortion on the prediction vectors.
The reason can be attributed to the fact that the link stealing attacks rely on, and are sensitive to, the similarity difference between adjacent nodes and non-adjacent nodes.
Our GRID can effectively reduce such differences with slight noise, for hiding the adjacent nodes into the indirect neighboring nodes to thwart the attacks.
In addition, we notice that when  $\theta$  increases to 0.4, the recall values drop to around $60\%$ for all attacks. 
Hence, a suggested $\theta$ value of 0.4 is well enough to defend against all link-stealing attacks.

\vspace{-0.5 em}
\subsubsection{The Impact of n-hop}

The $n$ value will affect the similarity gap in the objective function and the threshold $\delta$ for core node selection.
Here, we set the distortion budget $\theta=0.3$ and vary $n$ from 2 to 6, to show the link stealing attacks' recall values and the core node set size variations.
We conduct experiments for the Citesser dataset on GCN as the example, with the results shown in Fig.~\ref{Fig:N:citeseer}. 
We observe that when $n$ increases, the node amounts in the core node set rise marginally while the recall values appear to drop. 
For instance, when n increases from 2 to 6, the number of core nodes increases from $1515$ to $1569$, $1586$, $1594$, and $1596$.
The recall values corresponding to eight attacks drop  from $0.701$ to $0.621$, $0.642$ to $0.610$, $0.692$ to $0.653$, $0.711$ to $0.650$, $0.690$ to $0.631$, $0.689$ to $0.612$, $0.7311$ to $0.673$, and $0.681$ to $0.620$, respectively. 
The reason is that when $n$ is larger, there are fewer correlations for a node and its $n$-hop nodes, resulting in smaller values of ${\delta}$.
Hence, fewer links will be discarded in Algorithm~1, leading to an increased number of selected core nodes. 
Meanwhile, our GRID will enforce the similarity between adjacent nodes to the less correlated nodes, which thus can better degrade the attack performance. 
However, this requires generating larger similarity distortion, which may incur large distortion in prediction vectors.
Besides, as the results are shown in Table~\ref{Tab:core_2}, the increase in $n$ will also increase the computation time as more noises need to be calculated.
Hence, a careful selection of $n$ is required to balance defense performance, distortion loss, and computation time.

\vspace{-0.5em}
\subsection{Comparative Results}
\vspace{-0.5em}
\label{subsec:compare}
{We compare our approach with the prominent defense solutions against general inference attacks, as well as DP-based approaches specifically designed to defend against link-inference attacks.}

\subsubsection{Counterparts}
{Five state-of-the-art defense methods are taken as our counterparts, which are 
L2-Regularizer~\cite{shokri2017membership}, Dropout~\cite{srivastava2014dropout}, Min-Max Game~\cite{nasr2018machine}, MemGuard~\cite{jia2019memguard}, and the DP-based method, GAP~\cite{sajadmanesh2023gap}. }
Among them, L2-Regularizer, Dropout, Min-Max Game, and MemGuard were originally designed to defend against membership inference attacks.
{GAP was a novel DP-based GNN method, reported to achieve the best performance among DP-based defenses~\cite{sajadmanesh2023gap}}
The method details are illustrated in Appendix~\ref{sec:counter}.

\begin{figure*}
	\centering
        \subfigure[ GAP on GAT  with Cora  ]{
		\includegraphics[width=0.21\linewidth]{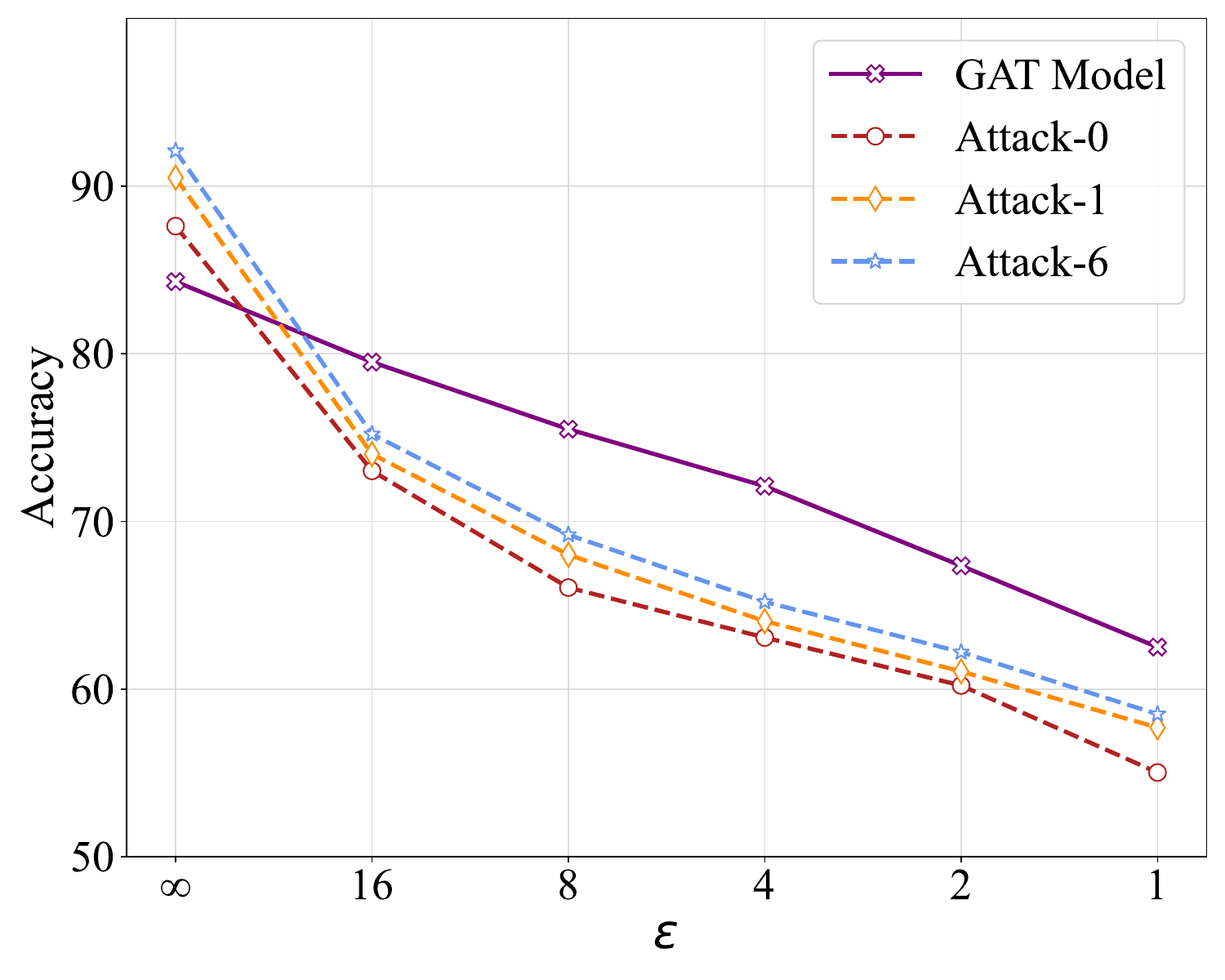}

	}	
	\subfigure[GRID on GAT  with Cora]{
		\includegraphics[width=0.21\linewidth]{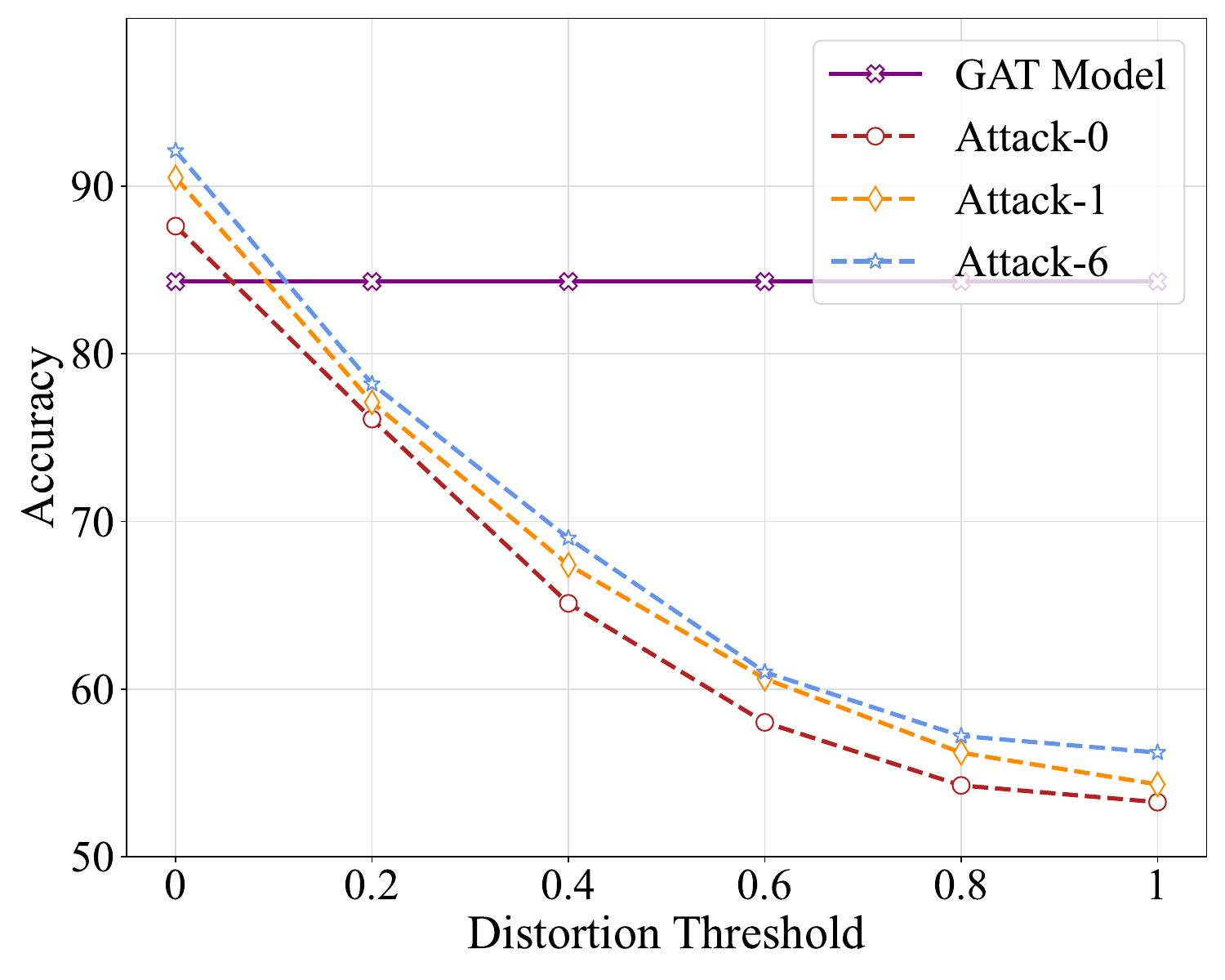}
	}
	\subfigure[GAP on GAT  with AIDS ]{
		\includegraphics[width=0.21\linewidth]{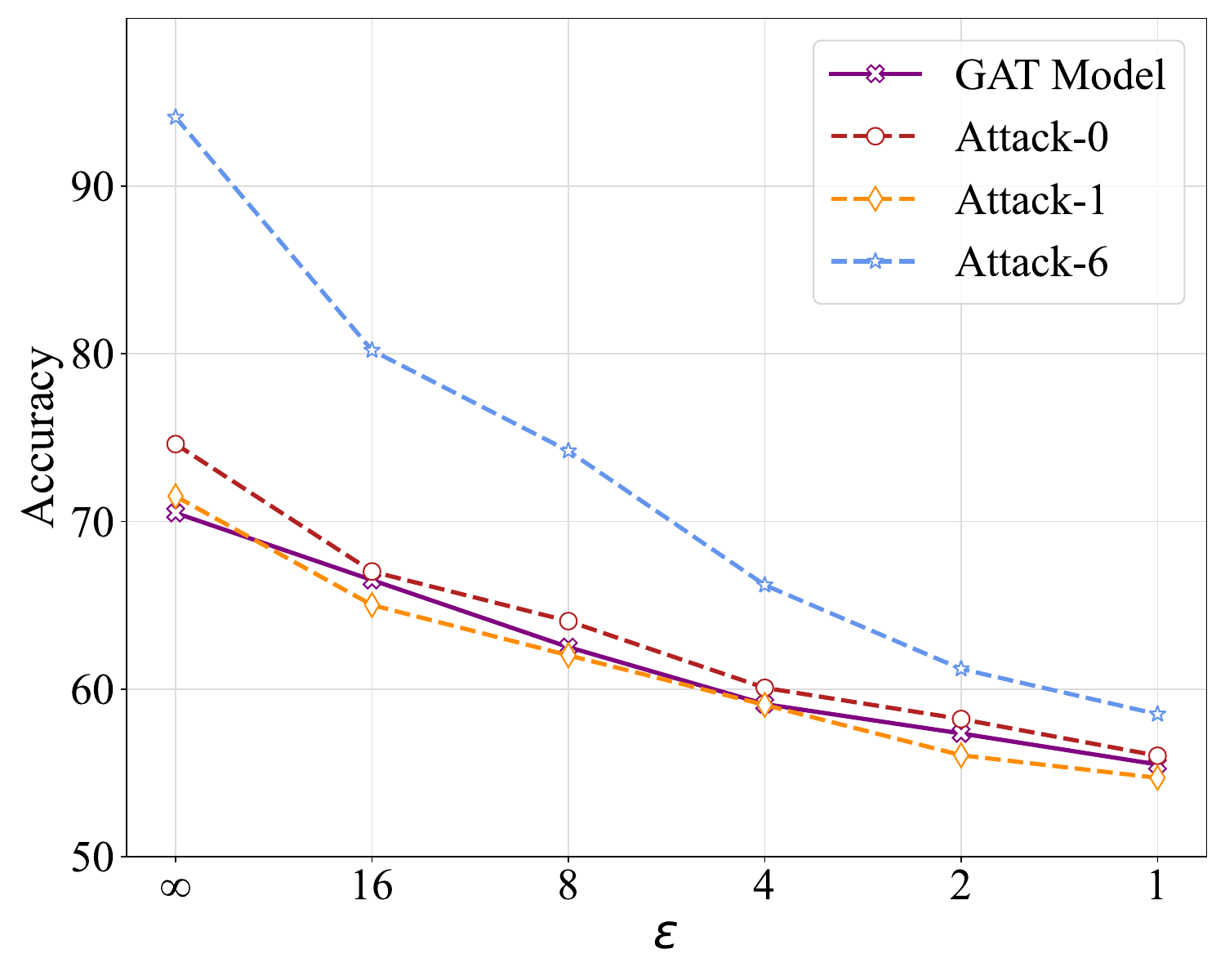}
	}	
	\subfigure[GRID on GAT  with  AIDS]{
		\includegraphics[width=0.21\linewidth]{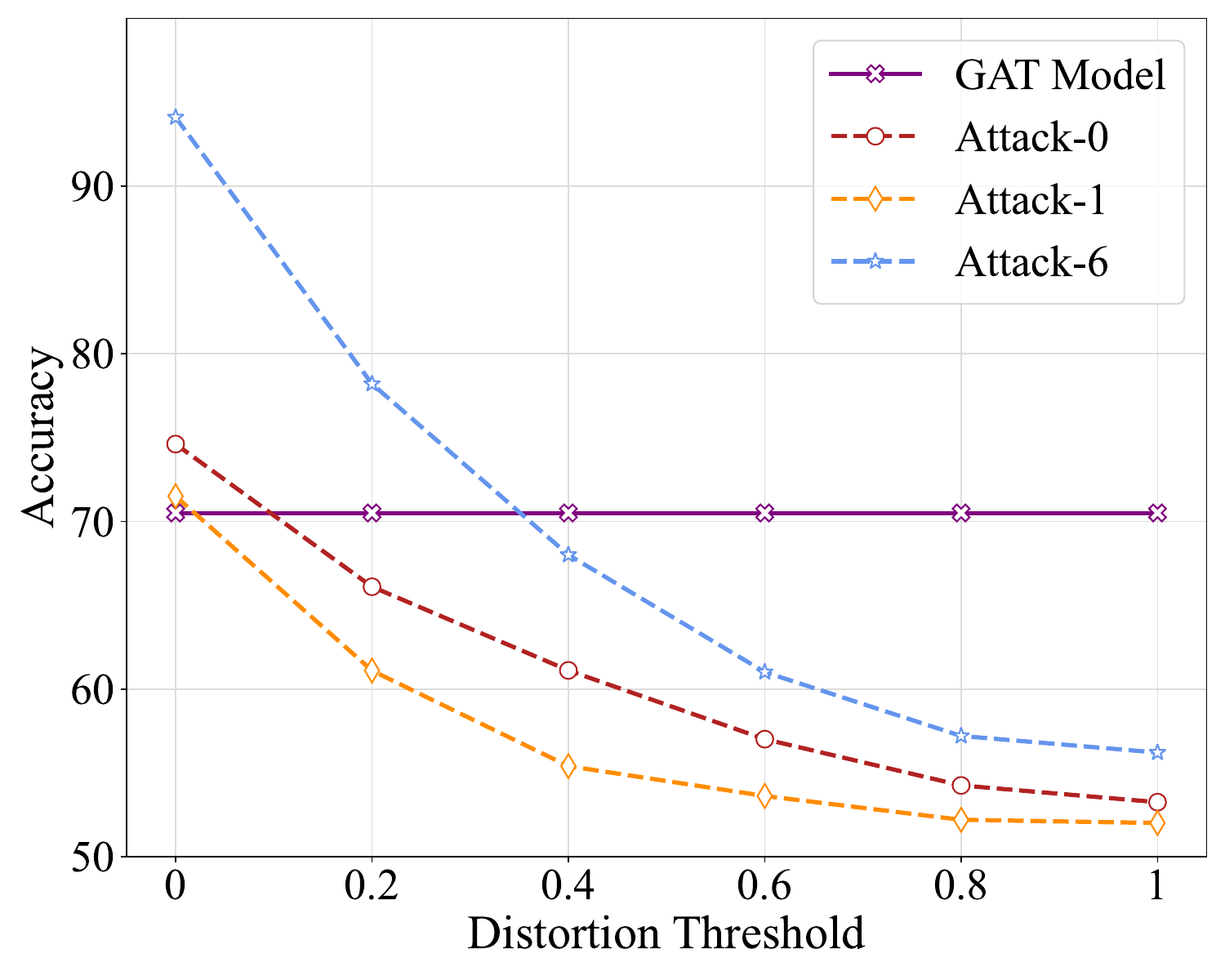}
	}
	\caption{The model prediction accuracy and link stealing attack accuracy variations for GAP and GRID with different hyperparameters.}
	\label{fig:GAP}
\end{figure*}

\begin{figure}
	\centering
	\vspace{-0.8em}
	\subfigure[ GAN-Citeseer for Attack-0  ]{
		\includegraphics[width=0.46\linewidth]{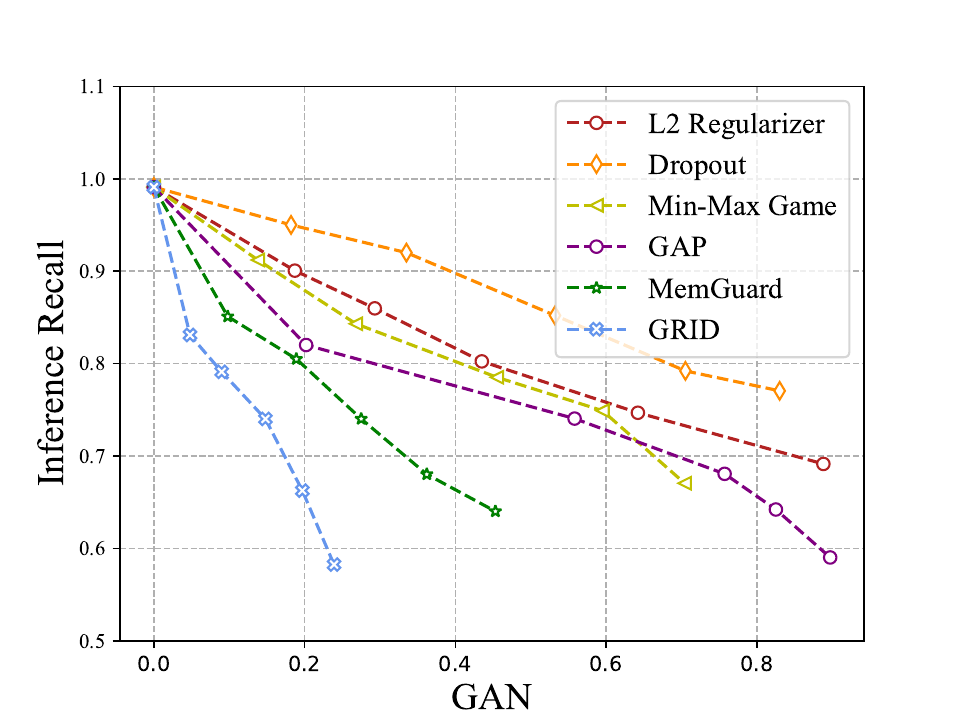}
		\label{Fig:GAN-0-citeseer}
	}\vspace{-0.8em}
	\subfigure[ALS-Citeseer for Attack-0]{
		\includegraphics[width=0.46\linewidth]{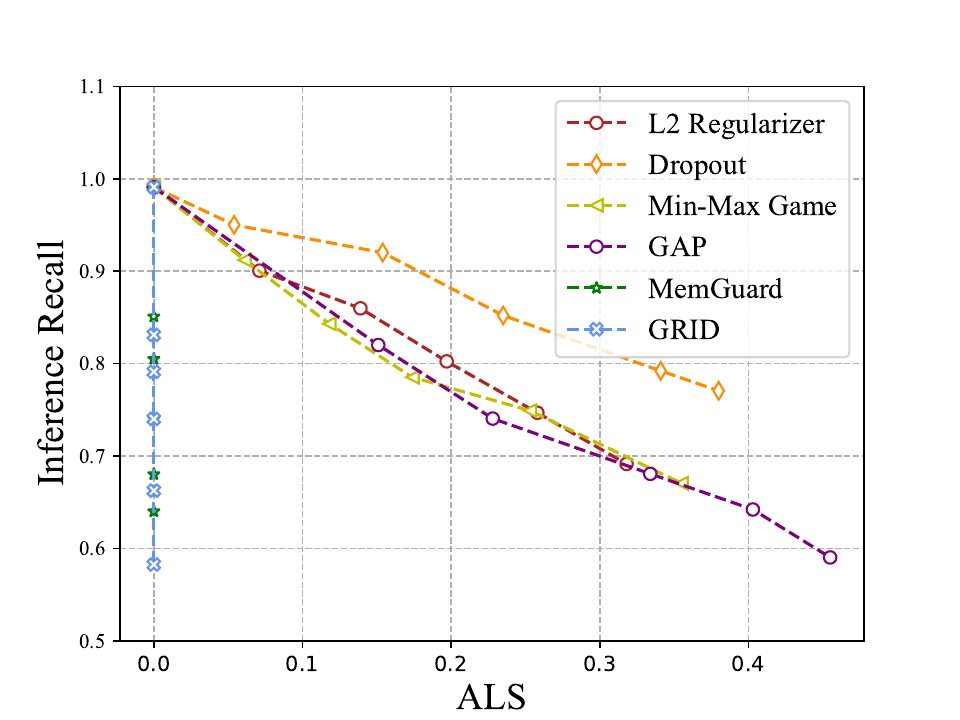}
		\label{Fig:ALS-0-citeseer}
	}
\vspace{-0.6em}
	\subfigure[ GAN-Citeseer for Attack-1 ]{
		\includegraphics[width=0.46\linewidth]{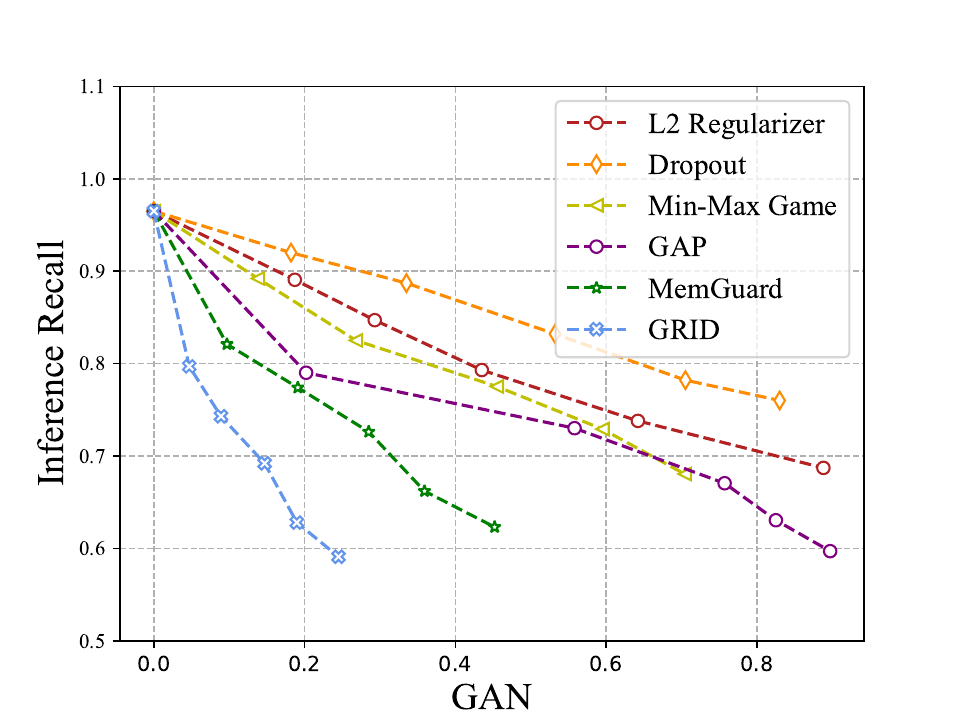}
		\label{Fig:GAN-1-citeseer}
	}
	\subfigure[ALS-Citeseer for Attack-1]{
		\includegraphics[width=0.46\linewidth]{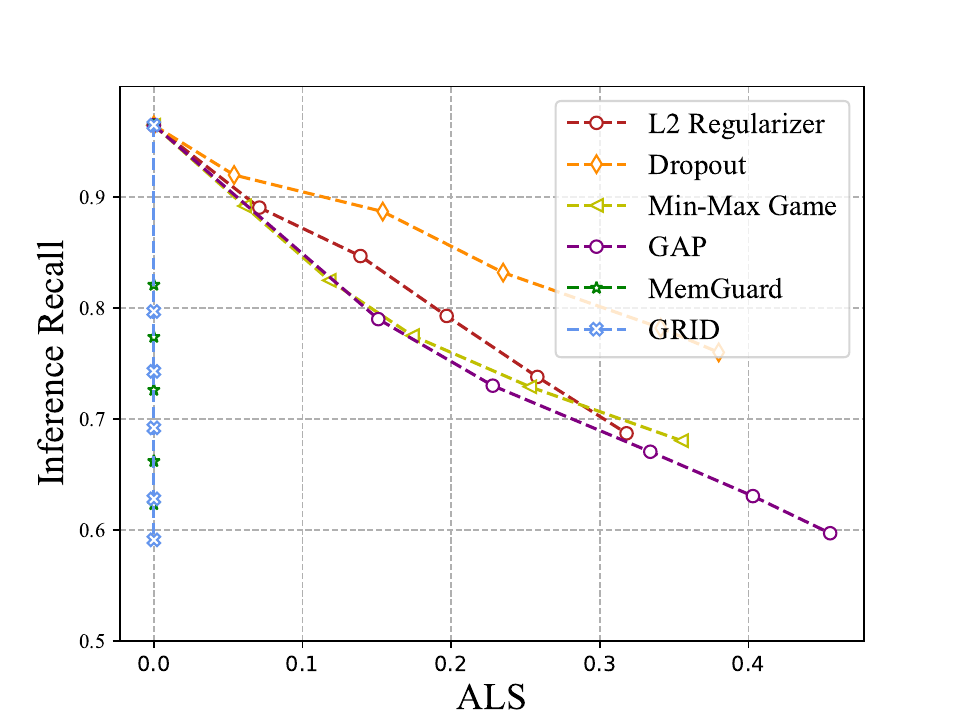}
		\label{Fig:ALS-1-citeseer}
	}
	\subfigure[ GAN-Citeseer for Attack-6 ]{
		\includegraphics[width=0.46\linewidth]{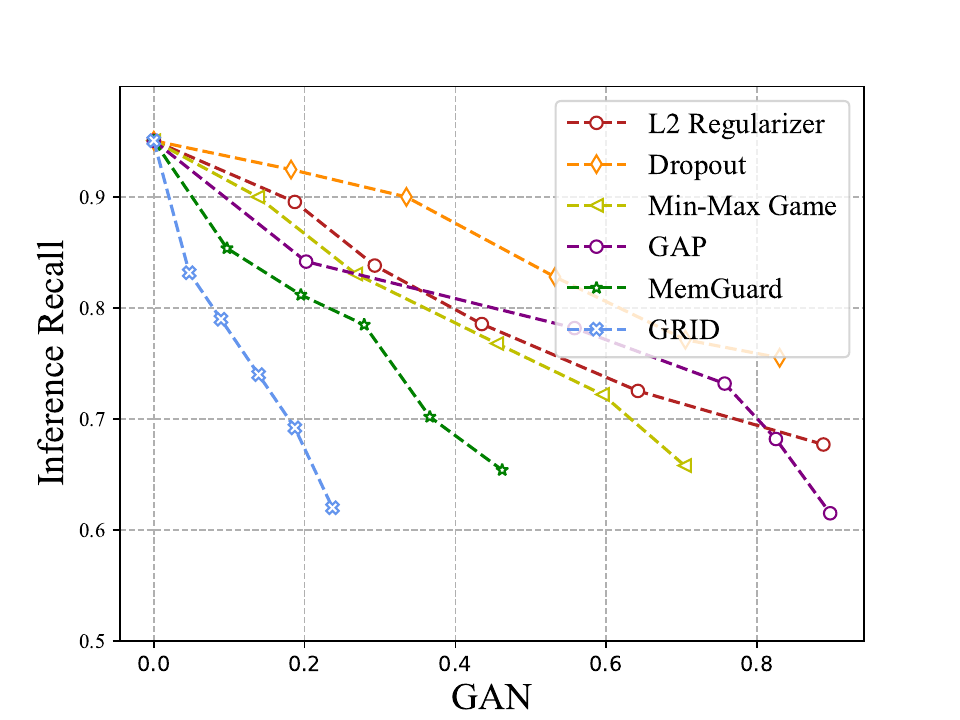}
		\label{Fig:GAN-6-citeseer}
	}
	\subfigure[ALS-Citeseer for Attack-6]{
		\includegraphics[width=0.46\linewidth]{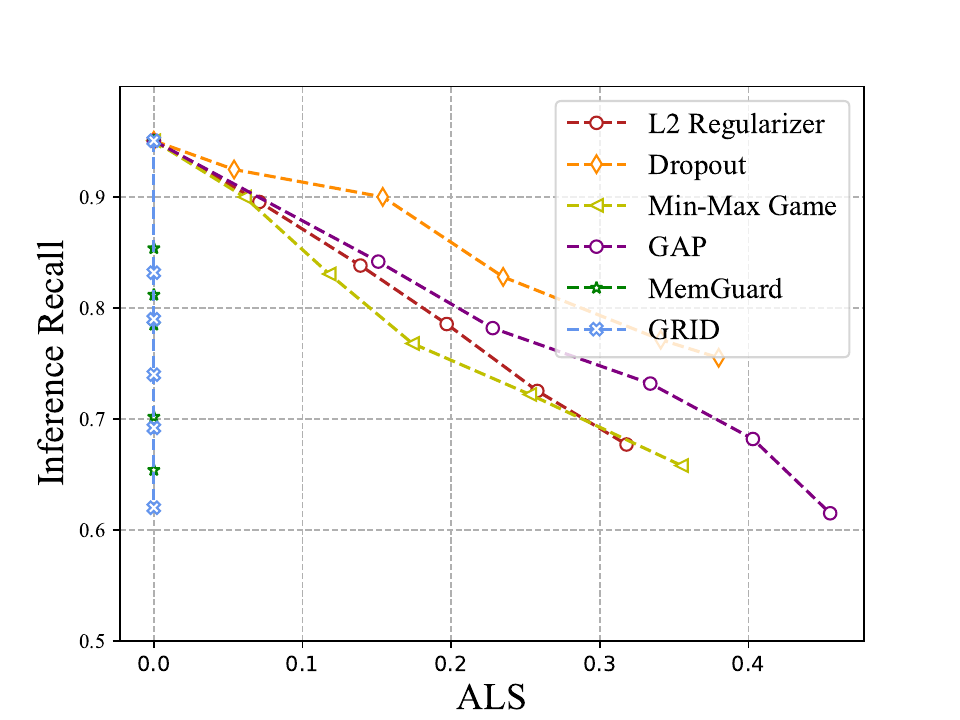}
		\label{Fig:ALS-6-citeseer}
	}
\vspace{-1em}
	\caption{Trade-offs between the defense performance and the utility loss, including graph-averaged noise and averaged label loss, for Attack-0, Attack-1, and Attack-6 on Citeseer.}
	\label{fig:attack-C}
	\vspace{-1.5em}
\end{figure}

\subsubsection{Defense Settings and Metrics}
To compare the performance of different defense methods, we will show the attack's recall-utility loss curve with the defense of each method, to demonstrate the trade-off between the defense performance and the utility loss.

\smallskip
\noindent{\bf \em Hyperparameter Configurations.}
For the L2-Regularizer, Dropout, Min-Max Game, and GAP, we first train the target model without defense to record the prediction vectors for some testing samples.
Then, we equip each defense under each hyperparameter to retrain the target model and record the new prediction vectors for the same testing samples.
By comparing the prediction vectors from the model with and without defense, we can characterize different utility losses on these prediction vectors.
Meanwhile, we perform the link-stealing attacks on the target model to obtain the corresponding attack recall values.
We record the pairs of attack recall and utility loss under different hyperparameter values to draw the curve.
For MemGuard, under each distortion budget, we calculate the noise vectors for all prediction vectors.
Then we perform the link-stealing attack on these distorted vectors and calculate the utility loss to obtain the pairs of attack accuracy and utility loss values. 
Similar to MemGuard, our GRID controls the distortion budget to calculate the noise vectors for the core nodes' prediction vectors and performs the link-stealing attack to obtain the pairs of attack recall and utility loss values. 

We sample six trade-off points under each defense method.
For L2-Regularizer, the hyperparameter is in the range [0, 0.05] with a step size of 0.01; for Min-Max Game, the hyperparameter is in the range [0, 2.5] with a step size of 0.5; and for Dropout, the hyperparameter is in the range [0, 0.8] with a step size of 0.15.
For GAP,  the hyperparameter $\epsilon$ is set as [$\infty,16,8,4,2,1$], while $\infty$ represents no defense.
For both MemGuard and our GRID, the distortion budget is in the range of [0, 0.5] with a step size of 0.1. 

\smallskip
\noindent{\bf \em Metrics.}
We consider two metrics to characterize the utility loss on the prediction vectors resulting from defense methods corresponding to each hyperparameter value.
The first one is   the graph-averaged noise ({\bf GAN}), which measures the average difference between the original prediction vector and the distorted prediction vector for all nodes in the training graph, calculated by

\begin{equation*}
	{\bf GAN}:\big(\sum_{n_i\in\mathcal{N}_c}d(v_i,v_i+s_i)\big)/{\vert\mathcal{N}\vert}\ ,
	\label{GAN}
\end{equation*}

where $\mathcal{N}$ is the node set,
Notably, {\bf GAN} characterizes the extent of distortions bringing to the target model.
The second one is the averaged label loss ({\bf ALS}), which is the fraction of the  nodes whose prediction labels, according to their prediction vectors, are altered after employing the defense mechanisms, defined as
\begin{equation*}
	{\bf ALS}:\Big(\sum_{n_i\in\mathcal{N}}\mathbbm{1}(\arg\max\{\mathbf{v_i}\},\arg\max\{\mathbf{v_i^*}\})\Big)/{\vert\mathcal{N}\vert}\ ,
	\label{LAS}
\end{equation*}
where $\mathcal{N}$ is the node set, $\mathbf{v_i}$ and $\mathbf{v_i^*}$ are the prediction vectors with and without defense, 
and $\mathbbm{1}()$ equals 1 if the two terms are the same and equals 0, otherwise.

\vspace{-0.5em}
\subsubsection{Comparative Results}
\label{sub:compare}
\vspace{-0.5em}
It's important to note that to cover five datasets, eight attacks, four models, and two metrics, a total of 320 groups of results would be needed. 
Thus, we only show some results from certain datasets and models while selecting Attack-0, Attack-1, and Attack-6 in~\cite{he2021stealing} as examples to provide comparative results.
Figure~\ref{fig:attack-C} presents the trade-off results between defense performance and the utility loss on Citeseer for GRID and its counterparts. 
From these figures, we observe GRID always achieves the best trade-offs under both attacks: with its curves consistently residing at the leftmost.

Figures~\ref{Fig:GAN-0-citeseer}, \ref{Fig:GAN-1-citeseer}, and \ref{Fig:GAN-6-citeseer}  show the trade-offs between defense performance and graph-averaged noise (GAN) on the Citseer dataset under GRID and its five counterparts.
Recall that GAN measures the total amount of noise distortion involved in the prediction vectors of all nodes in the training graph.
Taking Attack-1 in Figure~\ref{Fig:GAN-1-citeseer} as an example, five existing defense methods lower the attackers' recall values to below 70\% at GAN values of $0.868$, of $Null$ (dropout can't achieve this performance), of $0.704$, of $0.825$, and of $0.359$, respectively.
In contrast, our GRID achieves the same defense performance at the GAN value of only 0.147. 
This indicates that our GRID requires much less noise to be added to the target model in order achieve the same defense goal.
On the other hand, considering a smaller GAN value, say 0.2, our GRID lowers the recall value of Attack-1 to $62.80\%$ while other methods see their recall values to stand at $89.07\%$, $92.01\%$, $89.21\%$, $79.01\%$, and $82.01\%$, respectively.
Such results exhibit that GRID can achieve a much better defense performance at the same noise strength.
Clearly, GRID outperforms all existing defense mechanisms in terms of the trade-off between the defense goal and introduced noise.
The reason lies in the GRID strategy of adding noise vectors exclusively to the prediction vectors of core nodes (1569 out of 3323 for Citeseer). 
It significantly reduces the total noise amount incorporated in the model.  

Figures~\ref{Fig:ALS-0-citeseer}, and \ref{Fig:ALS-1-citeseer}, \ref{Fig:ALS-6-citeseer} draw the trade-off curves between defense performance and the averaged label loss under our GRID and five counterparts.
We observe that both our GRID and MemGuard achieve nil label loss, while the other three defense methods exhibit considerable label losses to achieve a given defense performance. 
The reason is that GRID and MemGuard regulate the label loss constraints to enforce that no prediction label is altered when adding noise vectors.
However, the other four methods will involve additional operations in the training process, i.e.,  calculating the regulation loss, turning off neurons, or adding noise to the data sample, which inevitably compromises their prediction accuracy.

{To better compare the privacy-utility tradeoff to DP-based methods, we plot the results showing the variations in model prediction accuracy and link-stealing attack accuracy on GAT under Cora and AIDS, as shown in Figure~\ref{fig:GAP}. 
We observe that, for GAP, a smaller $\epsilon$ value offers better defense against attacks but results in significant model accuracy degradation (dropping to 60\% at $\epsilon$=1).
In contrast, our GRID approach achieves comparable defense performance at a higher distortion threshold, while maintaining zero degradation in model accuracy. 
Both GRID and DP-based approaches contribute to the privacy-utility tradeoff but from different perspectives.
While DP-based methods provide formal privacy guarantees, they interfere with the training process or data, leading to inevitable declines in model performance. 
On the other hand, our GRID method provides a formal guarantee for model utility without affecting model training or data, although its privacy protection is based on empirical results.}

Hence, we conclude that our GRID mechanism achieves superior defense performance in GNNs while incurring the least utility losses than its counterparts examined.

\vspace{-0.5em}
\section{Discussion and Future Work}
\label{sec:discussion}
\vspace{-0.5em}

\begin{table*}[htb]
	\centering
	\caption{Influence-based link stealing attacks performance (\%) on GAT under five datasets with and without our GRID}
 \resizebox{0.9\linewidth}{!}{%
	\begin{tabular}{cl|cc|cc|cc|cc|cc}
		\toprule
		&& \multicolumn{2}{c|}{Citeseer}      & \multicolumn{2}{c|}{Cora}                & \multicolumn{2}{c|}{Pubmed} & \multicolumn{2}{c|}{AIDS} & \multicolumn{2}{c}{ENZYMES} \\
		&& No Defense & GRID                 & No Defense & GRID                       & No Defense      & GRID     & No Defense     & GRID    & No Defense      & GRID      \\
		\midrule
		\multirow{4}{*}{LinkTeller}& Accuracy & 83.65  & 66.03  & 84.76  & 67.84  & 82.07  & 63.08  & 73.54  & 59.58  & 71.25  & 58.20  \\
            & Precision & 83.63  &  {58.70}  & 83.36 &  {57.92}  & 82.24  &  {56.75}  & 73.05  &  {54.04}  & 71.74  &  {54.72}  \\
        & Recall   & 83.95  & 75.01  & 85.04  & 74.77  & 81.67 & 73.31  & 74.81  & 68.03  & 70.02  & 62.75  \\
        & AUC      & 84.70  &  {60.60}  & 83.55  &  {61.02}  & 81.79  &  {57.01}  & 73.57  &  {57.81}  & 72.18  &  {57.90} 
    \\
    \midrule
        \multirow{4}{*}{Link-Infiltrator}& Accuracy & 91.35  & 74.53  & 92.76  & 75.84  & 89.97  & 70.58  & 80.54  & 66.58  & 79.25  & 63.20  \\
		& Precision & 91.13  &  {64.20}  & 91.86 &  {64.42}  & 90.24  &  {61.25}  & 81.05  &  {59.54}  & 79.74  &  {56.05}  \\
		& Recall   & 91.45  & 83.51  & 93.54  & 83.27  & 89.67 & 81.81  & 82.31  & 76.53  & 78.52  & 69.25 \\
        & AUC      & 92.20  &  {66.10}  & 91.05  &  {66.52}  & 89.29  &  {65.51}  & 82.07  &  {64.31}  & 85.68  &  {64.40} \\
  \midrule
	\multicolumn{2}{c|}{Model Accuracy} & {75.80} &  {75.80} &  {85.60} &  {85.60} &  {83.30} &  {83.30} & {70.50} &  {70.50} &  {71.41} &{71.41}\\
		\bottomrule    
	\end{tabular}
 }
	\label{Tab:influence}
\end{table*}

\subsection{Defense against Influence-based Attacks}
\label{subsec:exp_influence}
In this paper, we mainly focus on the similarity-based link-stealing attack. But, we also explore our GRID performance in safeguarding the inductive GNN model from influence-based attacks.
Considering that influence-based attacks require attackers to inject nodes into the training graph and query the model to get predictions for these nodes, we accordingly tailor GRID to calculate the noises for the prediction vectors of injected nodes each time based on their current neighboring nodes.
That is, the attacker can only obtain noisy predictions for the injected nodes.

{We perform the latest attacks LinkTeller~\cite{wu2022linkteller} and Link-Infiltrator~\cite{meng2023devil} on the GAT model as  examples. 
For LinkTeller,  we follow their published code and set the hyperparameter k  as the accurate graph density to achieve optimal attack performance.
For Link-Infiltrator, the threshold is set as $e^{-7}$ as reported in~\cite{meng2023devil}.
For our GRID, we set the distortion budget $\theta=0.4$ and the hop $n=3$. 
The results are presented in Table~\ref{Tab:influence}.
From this, we observe that GRID's performance against influence-based attacks remains robust. 
Although the recall merely decreases by around 10\%, GRID effectively reduces attack precision and AUC to around 60\%. 
Notably, the low precision and AUC in link inference indicate significant misclassification of links between non-neighboring nodes, such as incorrectly identifying connections between individuals who are not acquainted in a social network. 
The reason is that our GRID method introduces noise to the prediction vectors of the query nodes before and after injecting nodes, consistently causing random variations in their prediction vectors. 
This misleads attackers into interpreting such variations as the influence of injected nodes transmitted through the links, leading them to predict false positives, thereby degrading the attack's precision.
On the other hand, such noise may also cancel the influence caused by the injection node even if two target nodes are linked, causing false negative predictions.
However, it is not guaranteed as evidenced by the minor degradation in the attack recall.
In our future work, we will consider incorporating massage passing influences into our noise generation tasks to further enhance GRID's performance against influence-based attacks.}

\subsection{Defense against Adaptive Attacks}
\vspace{-0.5em}
In our design of GRID, we have proactively considered the potential scenario that an attacker may know our GRID mechanism and attempt to launch the adaptive attack in that it predicts the presence of a link between two nodes based on the low similarity of prediction vectors of adjacent nodes. 
To counter such adaptive attacks, our design aims at disguising adjacent nodes into the n-hop indirect neighboring nodes, hence, effectively thwarting such adaptive attacks. 
Here, we consider another adaptive attack by assuming that the attacker has the knowledge of the certain value of the $n$-hop.
In this attack, an attacker treats all n-hop indirect neighboring nodes as adjacent nodes when constructing the ground truth to train the link-stealing classifier. 
This allows attackers to learn the similarity patterns of adjacent nodes more effectively.
To assess the resilience of our GRID against this adaptive scheme, we utilized Attack-6, which represents the strongest attack, as the backbone for implementing this adaptive strategy. 
Here, we set the distortion loss $\theta=0.4$ and $n=3$ for our GRID. 

\begin{table}[t]
	\centering
	\scriptsize
	\caption{Adaptive  attack performance under our GRID}
	\vspace{-0.5em}
	\begin{tabular}{c|ccccc}
		\toprule
		& Citeseer & Cora & Pubmed & AIDS & ENZYMES \\
		\midrule
		Precision &  0.591        &    0.598  &     0.572   &  0.589    &  0.575       \\
		Recall    &    0.772      &  0.763    &    0.774    &  0.723    &   0.715      \\
		Accuracy    &    0.682      &  0.681    &   0.672     &  0.657    &     0.647   \\
		AUC       &     0.792     &   0.793   &  0.774      &  0.723    &     0.715   \\
		\bottomrule
	\end{tabular}
	\label{Tab:Adaptive}
	\vspace{-1.5em}
\end{table}

Table~\ref{Tab:Adaptive} lists our results on five datasets. 
Comparing to Table~\ref{Tab:attack-6}, we observe that such an adaptive attack has a positive impact on the recall values, boosting them to approximately $75\%$ due to its ability to learn additional similarity patterns of adjacent nodes, thereby improving link recognition.
However, it also comes at the cost of deteriorating the attack precision to lower than $60\%$. 
This decline occurs since the adaptive attack wrongly identifies n-hop indirect neighboring nodes as adjacent nodes when constructing the ground truth for training the link-stealing classifier. 
The trade-off between inference recall and precision underscores the robustness of our preventive measure, which disguises the similarity levels of adjacent nodes into the n-hop indirect neighboring nodes. 
By doing so, GRID can effectively hinder attackers from exploiting such adaptive strategies. 
However, this attack is merely an example of the heuristic argument; in practice, the attackers may devise other strong adaptive attacks, such as optimization problem-based schemes.
Thus, in our future work, we will refine defense formulations and provide a mathematical proof of its guaranteed performance under stronger adaptive attacks.

\section{Related Work}
\label{sec:related}

\smallskip
\noindent{\bf Link Inference Attack.}
The link inference attacks toward GNN models were proposed in~\cite{he2021stealing}, aiming at stealing the link of the training graph. 
It leverages the characteristics of the GNN model, which aggregates information for each node from its adjacent nodes, causing a high similarity in the prediction vectors of adjacent nodes. 
Hence, an attacker can measure the similarity pattern of two queried nodes to determine whether a link exists between them. 
Subsequent work employs such a similarity-based method to attack the unsupervised graph representation learning or inductive GNN models~\cite{wang2023link,wu2024link}. 
Besides,~\cite{wu2022linkteller, meng2023devil} considered the scenario where the attacker can inject the malicious nodes around the target nodes in the training graph.
Then, the attacker can measure the variation of prediction vectors queried for the target node before and after node injection to infer whether they are linked, which is called the influence-based attack.

\smallskip
\noindent{\bf Membership Inference Attack}
The membership inference attacks aim to infer if a data point belongs to the training dataset of deep learning. 
It was first proposed in~\cite{shokri2017membership}, and then  
plentiful follow-up research was conducted to improve the original designs or apply them to different learning algorithms/models~\cite{salem2019ml,li2021membership,leino2020stolen, choquette2021label,liu2021encodermi,salem2020updates,nasr2019comprehensive,sablayrolles2019white,song2021systematic,song2019privacy,hidano2021transmia,carlini2021extracting,song2020information,li2021membership2}, such as contrastive learning, online learning, transfer learning, federated learning, natural language domains, among others. 
These attacks are not designed originally for GNN models, so their extension to steal links in the training graph of a GNN remains open.
On the other hand, some studies~\cite{he2021node,wu2021adapting,olatunji2021membership} have extended the membership inference attack into the context of GNNs, for inferring whether a node sample was in the training graph.
However, the exploration of defending membership inference attacks is out of the scope of our work.

\smallskip
\noindent{\bf Model Inference Attack.}
The model inference attacks have been proposed for stealing model parameters 
in~\cite{chandrasekaran2020exploring,jagielski2020high,tramer2016stealing,wang2018stealing}.
The attacker typically doesn't have direct access to the target model's architecture, weights, or training data. 
Instead, they query the model with selected inputs and observe the corresponding outputs or predictions to infer the model parameters for replicating the functionality of a model.
Since we focus on the link inference attacks toward the training graph, this line of attacks is not under consideration in the current scope.

\smallskip
\noindent{\bf Defense for Inference attacks. }
Some defense methods have been proposed to defend inference attacks and they can be roughly grouped into three categories: 
1) by employing differential privacy~\cite{shokri2015privacy,abadi2016deep,wang2017differentially,yu2019differentially,wu2022linkteller,sajadmanesh2023gap}; 2) by adding regularization terms in the loss function~\cite{li2021membership2,nasr2018machine,salem2019ml,song2021systematic}; 3) by adding noise in the model predictions~\cite{jia2019memguard}.
The first two categories of solutions inevitably disrupt the training process of a target model, thus incurring considerable model accuracy degradation.
For example, the L2-Regularizer~\cite{shokri2017membership} and Min-Max Game~\cite{nasr2018machine} added additional terms for overfitting prevention, inevitably degrading the accuracy of the target model.
The differential privacy methods suffer considerable utility loss, as presented in the pioneering work~\cite{sajadmanesh2022gap}, where a target GNN model has an over 20\% degradation in model accuracy when defending against the node inference attack. 
On the other hand, MemGuard~\cite{jia2019memguard} was proposed to defend against membership inference attacks,
by adding noise vectors to the prediction vectors of a trained model for distortion, making it difficult for an attacker to infer if a queried data sample belongs to the member of the training dataset. 
The utility loss of the prediction vector is kept within bounds by moderating the intensity of the noise vectors.
However, the protection of MemGuard for the GNN model is limited, as the noise is computed for distorting the relationship of the individual node prediction vector and its attributes, without taking the correlations among prediction vectors into consideration. 
Hence, it cannot capture the graph topology structure information when crafting the noises.

\section{Conclusion}
\label{sec:conclusion}
 
{This paper has presented a novel defense solution for GNN models, called GRID, for countering link stealing attacks with the model utility guaranteed formally.}
Our GRID takes into account the graph topology to craft noise vectors for core nodes in the training graph to disguise the similarity of prediction vectors of adjacent nodes as those of n-hop indirectly connected neighboring nodes. 
We formulate the noise generation objective into an optimization problem and propose an algorithm for identifying a subset of nodes as core nodes for calculating noise vectors. 
{The extensive experimental results on different datasets exhibit that GRID can effectively defend against different types of link-stealing attacks and outperform its counterparts in terms of defense performance and model utility tradeoffs.}

\section*{Acknowledgments}
This work was supported in part by NSF under Grants 2019511,
2348452, 2315613, 1937787, and 2325564. 
Any opinions and findings expressed in the paper are those of the authors and do not necessarily reflect the views of funding agencies.

\bibliographystyle{plain}
\bibliography{main}

\appendices 

\section{Additional Calculation Details and Results}
\label{sec:appendix}

\subsection{Similarity Calculations}
\label{sec:SimCal}

$corr()$ and $cos()$ measure the Pearson correlation coefficient and cosine metrics between two vectors $\mathbf{v}_i $ and $\mathbf{v}_j$, which is calculated as
\begin{align*}
	corr({\bf X},{\bf Y})&=\frac{\sum((x_i - \overline{X})(y_i - \overline{Y}))}{\sqrt{\sum(x_i - \overline{X})^2 \sum(y_i - \overline{Y})^2}}\notag \\
	cos({\bf X},{\bf Y})& =\frac{\sum_{i=1}^{n} x_i \cdot y_i}{\sqrt{\sum_{i=1}^{n} x_i^2} \cdot \sqrt{\sum_{i=1}^{n} y_i^2}}
\end{align*}
where $\bar X$ and $\bar Y$ are the average values of ${\bf X}$ and of ${\bf Y}$.


\begin{table*}
	\centering
	\caption{Link stealing Attacks-0, -1, and -6 performance (\%) on GraphSAGE and GIN under five datasets with and without our GRID. Note: `a/b' are the values under GraphSAGE and GIN}
 \resizebox{\linewidth}{!}{%
	\begin{tabular}{cl|cc|cc|cc|cc|cc}
		\toprule
		&& \multicolumn{2}{c|}{Citeseer}      & \multicolumn{2}{c|}{Cora}                & \multicolumn{2}{c|}{Pubmed} & \multicolumn{2}{c|}{AIDS} & \multicolumn{2}{c}{ENZYMES} \\
		&& No Defense & GRID                 & No Defense & GRID                       & No Defense      & GRID     & No Defense     & GRID    & No Defense      & GRID      \\
		\midrule
		\multirow{4}{*}{Attack-0}& Accuracy & 85.8/85.5 & 65.1/64.7 & 84.4/84.1 & 62.2/61.9 & 77.6/77.3 & 60.6/60.4 & 74.0/73.9 & 59.4/57.1 & 73.1/72.6 & \textbf{56.9/56.6} \\
		& Precision & 76.6/76.2 & 65.4/64.9 & 75.6/75.4 & 61.7/62.3 & 68.4/68.2 & 56.4/56.0 & 51.4/55.3 & \textbf{50.2/51.3} & 52.5/52.2 & 50.6/52.4 \\
		& Recall   & 97.1/96.7 & 63.7/63.4 & 94.9/94.6 & 62.5/62.0 & 93.5/93.2 & 62.1/61.7 & 96.7/96.2 & \textbf{62.1/63.4} & 97.7/97.4 & 62.5/62.0 \\
        & AUC      & 90.0/90.9 & 68.1/67.8 & 85.8/84.6 & 67.5/67.1 & 85.4/83.7 & 66.7/66.3 & 69.5/68.9 & 58.5/58.1 & 62.2/61.8 & \textbf{55.2/55.0} \\ 
        \midrule
        \multirow{4}{*}{Attack-1}& Accuracy & 88.7/88.9 & 66.6/65.8 & 86.1/86.4 & 65.1/64.3 & 83.9/84.3 & 61.6/61.2 & 70.4/70.2 & 54.1/54.2 & 68.1/68.2 & \textbf{52.3/52.4} \\
        & Precision & 85.9/86.1 & 66.0/66.3 & 84.1/84.3 & 65.7/65.5 & 76.5/76.4 & 61.3/60.2 & 71.3/71.0 & 54.3/54.0 & 67.1/67.3 & \textbf{53.2/53.1} \\
        & Recall   & 94.6/94.8 & 67.4/67.2 & 87.1/87.3 & 64.8/65.0 & 88.0/88.2 & 61.5/62.4 & 69.2/69.4 & 53.5/51.3 & 69.1/69.2 & \textbf{51.8/52.0} \\
        & AUC     & 91.3/95.5 & 72.3/72.4 & 87.9/93.0 & 71.8/71.9 & 87.3/87.5 & 69.6/69.4 & 71.7/71.6 & \textbf{57.8/57.7} & 73.3/73.2 & 59.0/56.1 \\
        \midrule
        \multirow{4}{*}{Attack-6}& Accuracy & 91.4/90.9 & 67.2/66.9 & 92.2/89.7 & 67.6/67.1 & 90.2/89.7 & 68.0/66.4 & 93.4/92.8 & 67.5/63.4 & 81.6/81.2 & \textbf{64.3/63.6} \\
        & Precision & 89.3/88.7 & 65.4/66.8 & 93.4/90.4 & 67.9/67.2 & 89.5/88.8 & 68.9/67.3 & 89.9/89.3 & 68.7/66.1 & 76.3/75.7 & \textbf{64.6/64.0} \\
        & Recall   & 92.5/92.0 & 68.3/67.0 & 92.0/89.6 & 68.5/66.1 & 91.6/91.0 & 67.8/66.2 & 97.8/97.2 & 66.9/62.3 & 87.9/87.2 & \textbf{64.1/63.5} \\
        & AUC      & 97.3/96.8 & 70.4/68.8 & 95.6/95.0 & 69.7/67.1 & 96.2/95.7 & 70.1/69.4 & 97.1/96.5 & 73.1/68.8 & 88.4/87.7 & \textbf{67.4/66.9} \\

        \midrule
        \multicolumn{2}{c|}{Model Accuracy} & {74.8/72.7}  & {74.8/72.7}  & {77.6/76.2}  & {77.6/76.2}  & {85.4/82.3}  & {85.4/82.3}  & {70.2/72.5}  & {70.2/72.5}  & {71.1/71.9}  & {71.1/71.9} \\

		\bottomrule    
	\end{tabular}
 }
	\label{Tab:combined}
\end{table*}

\subsection{Gradient Calculation}
\label{sec:app:grad}
{Here, the gradient $\nabla_s\mathcal{L}$ on the vector ${\bf s}$ is calculated on each entry.
For $\lambda(v^a+s^a-v^*-s^*)$, we construct a vector ${\bf C}$,  which is the same length as $s$, with elements defined as
\begin{itemize}
    \item If corresponding to $s^*$, then $C_i=-1$.
    \item If corresponding to $s^a$, then $C_i=1$.
    \item Otherwise, $C_i=0$
\end{itemize}
Then, we have 
\begin{equation}
    \nabla_s[\lambda(v^a+s^a-v^*-s^*)]=\lambda\nabla_s(C^Ts)=\lambda C\ .
\end{equation}
For $\mu\sum_{a}(s^a)$, we have
\begin{equation}
    \nabla_s[\mu(\sum_{a}s^a)]=\mu\nabla_s(\textbf{1}^T\mathbf{s})=\mu \textbf{1}^T\ ,
\end{equation}
where $\textbf{1}$ is the vector of 1 as the same length as $s$.
For $\nu(\| \mathbf{s}\|-\theta)$, we have 
\begin{equation}
    \nabla_s[\nu(\| \mathbf{s}\|-\theta)]=\nu\nabla_s\| \mathbf{s}\|=\nu \frac{s}{\| \mathbf{s}\|}\ .
\end{equation}
Combine them together, we have:
\begin{equation}
    \nabla_s \mathcal{L} = \nabla_s D_i + \lambda C + \mu \textbf{1}^T+\nu \frac{s}{\| \mathbf{s}\|}\ .
\end{equation}}

\subsection{Counterpart Defense Methods}
\label{sec:counter}
\smallskip
\noindent{\bf L2-Regularizer.}
The first proposed membership inference attack~\cite{shokri2017membership} dealt with overfitting to lead to its success.
It is also considered an elementary defense method by employing the L2-Regularizer to prevent overfitting.
The main idea of  L2-Regularizer is to add a penalty term, by calculating the sum of the L2-norm value of the weights in the target model, as the loss function.

\smallskip
\noindent{\bf Dropout.}
Dropout~\cite{srivastava2014dropout} is considered as another method to prevent the model overfitting, by randomly turning off some neurons (i.e., setting their outputs to 0) during each training iteration of the neural network.
This way prevents the model from being sensitive to some particular neurons since any at-risk neuron is turned off at random to make its weight value as average as possible. 

\smallskip
\noindent{\bf Min-Max Game.}
Inspired by the L2-Regularizer, authors in~\cite{nasr2018machine} proposed the Min-Max Game method, which added a special adversarial regularizer to defend against the membership inference attacks.
This regularizer aims at maximizing membership privacy by narrowing the difference between the prediction vectors of the model on its training set and on other data samples with the same underlying distribution.

\smallskip
\noindent{\bf GAP.}
GAP was proposed in~\cite{sajadmanesh2023gap} as the latest and novel differentially private GNN based on aggregation perturbation rather than training graph perturbation.
By adding stochastic noise to the aggregation function of a GNN, GAP statistically obfuscates the presence of a single edge (edge-level privacy) or a single node and all its
adjacent edges (node-level privacy). 
GAP has a parameter privacy budget, i.e., $\epsilon$, to tune the privacy-utility trade-off, where a lower privacy budget leads to stronger privacy guarantees but reduced utility.

\smallskip
\noindent{\bf MemGuard.}
MemGuard was proposed in~\cite{jia2019memguard} to defend the membership inference attack by randomly adding noise to the prediction vectors of a target ML model.
The noise vector aims to counter random guessing results for the membership inference classifier by removing the membership pattern on the prediction vector that is related to the corresponding query sample. 
We employ the MemGuard to add noise vectors to the prediction vectors of all nodes in the training graph and control the distortion budget to evaluate its defense performance.

\subsection{Defense Performance on GraphSAGE and GIN}
\label{subsec:other_results}
{
The defense performance for attack-0, attack-1, attack-6 on GraphSAGE and GIN are shown in Table~\ref{Tab:combined}.
Since our GRID is independent of specific GNN structure designs, it consistently performs well on GraphSage and GIN. 
For example, GRID decreases the accuracy(\%) of the strongest Attack-6 to {67.2, 67.6, 68.0, 67.5, 64.3} and {66.9, 67.1, 66.4, 63.4, 63.6} for GraphSAGE and GIN on five datasets.}

\end{document}